\documentclass{article}
\usepackage[preprint]{tmlr}
\usepackage[utf8]{inputenc}
\usepackage[T1]{fontenc}
\usepackage{amsmath,amssymb}
\usepackage{booktabs}
\usepackage{graphicx}
\usepackage{xcolor}
\usepackage{url}
\usepackage{hyperref}
\usepackage{caption}
\usepackage[all]{hypcap}
\usepackage{wrapfig}
\usepackage{float}
\usepackage{longtable}
\usepackage{microtype}

\title{QuoteBench: How Matched Scores Can Hide Command-Path Failures}

\author{%
\centering
Shangao Li$^{1}$ \qquad
Yao Zhang$^{2,3}$\thanks{Yao Zhang and Yuanyuan Yang are corresponding authors.} \qquad
Volker Tresp$^{2,3}$ \qquad
Yuanyuan Yang$^{1}$\footnotemark[1]\\[0.35em]
{\normalfont\small
$^{1}$Stony Brook University \qquad
$^{2}$LMU Munich \qquad
$^{3}$Munich Center for Machine Learning}\\[0.25em]
{\normalfont\small
\href{mailto:shangao.li@stonybrook.edu}{shangao.li@stonybrook.edu} \qquad
\href{mailto:yaoz720.ai@gmail.com}{yaoz720.ai@gmail.com}\\[0.05em]
\href{mailto:tresp@dbs.ifi.lmu.de}{tresp@dbs.ifi.lmu.de} \qquad
\href{mailto:yuanyuan.yang@stonybrook.edu}{yuanyuan.yang@stonybrook.edu}\\[0.18em]
\href{https://quotebench.lsamc.website/}{Project page: quotebench.lsamc.website}}
}

\hypersetup{
  hidelinks,
  pdftitle={QuoteBench: How Matched Scores Can Hide Command-Path Failures},
  pdfauthor={Shangao Li, Yao Zhang, Volker Tresp, Yuanyuan Yang}
}

\begin{document}
\raggedbottom
\maketitle

\begin{abstract}
LLM coding agents issue Bash commands through interfaces that may serialize, wrap,
and reparse model output. Matched execution scores alone cannot distinguish command-
generation errors from failures introduced after generation. \textbf{QuoteBench}
measures this boundary with exact final-state validation on 56 one-shot tasks from
14 incident-derived families, crossing the generation contract with the execution
transport around one deliberately unescaped added parser. Escaping at the
interpolation point reproduces each replayed reply's raw-path outcome, so any
recovery under a disclosed boundary must come from the model changing its
generation. Across eight
same-window configurations, replaying the same reply through the added parser
lowers success by 55.4--73.2 points; disclosure recovers 30.4--60.7 points for six
configurations, and zero or slightly negative for the other two. Raw generation is nearly saturated at the
frontier; boundary adaptation is what still separates models: GPT-5.6-sol's
matched gap of $-3.6$ points hides $-64.3$ damage and $+60.7$ compensation. The
deployment configuration reorders models: one reversal among 26 comparable pairs
is unambiguous and four more sit on single-task margins. Evaluations of
command-issuing agents should report the model configuration, generation contract,
execution path, operating point, and final-state validator rather than treat a
matched score as an intrinsic model property.
\end{abstract}

\section{Introduction}

Bash quoting failures can corrupt literals, break routine agent actions, and trigger
repair loops. Even simple tasks such as writing exact bytes, passing a literal
argument, editing JSON, or invoking a remote-like wrapper must preserve quotes,
dollar signs, backticks, newlines, glob characters, and expansion timing
\citep{bashmanual2025,shellcheck,wheeler2010filenames}. Public issue trackers document
broken heredocs (the shell's inline multiline-string syntax), over-quoted operators, and repeated repair attempts
\citep{claudecodeheredoc2026,codexcontract2026,warpheredoc2025}. The full survey
appears in Appendix~\ref{app:coverage}.

A failure also consumes a model generation and tool invocation, while diagnosis and
retry remain in the trace. A common recovery is to write and execute a temporary
script, adding actions and possibly workspace artifacts. The incident evidence guides
family selection and mechanism coverage, not prevalence estimation.

Current benchmarks leave this failure mode entangled with other capabilities. Broad
coding and terminal benchmarks combine command construction with planning,
repository navigation, and recovery
\citep{agentbench2024,webarena2023,osworld2024,intercode2023,swebench2024,
terminalbench2026}, whereas command-generation benchmarks score emitted programs
under fixed transport \citep{nl2bash2018,nlc2cmd2021,nl2shalfa2025,bashcoder2026}.
Thus agent success does not reveal whether the first command preserved its payload,
and command-generation success does not show whether it survives deployment. Bash
quoting permits a focused test because parser rules are public and final state is
exactly checkable \citep{bashmanual2025}.

We introduce \textbf{QuoteBench}, a benchmark of one-shot LLM-generated Bash commands.
Its 56 tasks cover 14 operation families, each with one benign control and three
hazardous payload variants. The tasks exercise multiline text, hostile filenames,
regular-expression and glob metacharacters, heredocs, literal argv and environment
values, Git metadata, and two local SSH-like simulations. Validators inspect final
bytes, argv, JSON, directory state, or Git history, so any command that reaches the
correct state receives credit.

Agent interfaces range from direct shell actions to structured action languages
\citep{sweagent2024,codeact2024,kim-etal-2026-codestruct}. We call a reply that runs
as the shell program \emph{raw}, and a command field extracted from a structured tool
call \emph{native}. A fixed-commit survey of six public agent systems finds both in
use (Appendix Table~\ref{tab:harness-survey}).

Commands targeting remote or containerized machines can cross another boundary and
be reparsed inside double quotes, for example through \texttt{ssh host "\ldots"},
\texttt{docker exec sh -c "\ldots"}, or a CI \texttt{run:} step. QuoteBench uses this pattern as the
\emph{nested} condition, a controlled intervention that adds one downstream parser.
Five of the seventeen retained public incidents contain such a boundary.

We ask how reliability varies across models, operation families, and command paths,
how generation contract interacts with execution transport for fixed replies, and
what provider-exposed effort ladders reveal about matched and replayed behavior.
At the fixed configurations used for the same-window crossover, matched nested
success spans 14.3--91.1\%. Table~\ref{tab:scorecard} separately selects each model's
best observed measured operating point, where three models reach 100.0\% and the
remaining scores span 14.3--98.2\%. Separately, six provider-hosted models using native shell
tools score 85.7--98.0\%. Effort improves matched success for some models but not
others, and the same effort label corresponds to different token budgets across
models.

To separate generation errors from transport damage, we replay each fixed
raw-conditioned reply with and without one added double-quoted parser. This intervention
reduces success by 55.4--73.2 points in every same-window configuration. Under the same transport, contract-conditioned generations recover 30.4--60.7
points for six of eight configurations. This realized contrast is computed over the
stored generations in the frozen benchmark. Across the observed trial-0 effort rungs, the
unconditional nested-replay pass rate varies by at most 5.4 points within each
measured ladder. The same mechanism persists on
private payloads and across repeated draws.

Concurrent work already shows that changing the harness reorders model
leaderboards~\citep{harnessdisclosure2026}, but because it swaps the whole scaffold it measures variance without
attributing a reversal to any one mechanism. By fixing the model output and changing
a single parser, QuoteBench attributes the reorder to the command path and
decomposes the matched score into transport damage and contract-conditioned
compensation.

This paper makes three contributions:
\begin{enumerate}
  \item \textbf{A final-state benchmark of command-path reliability.} QuoteBench
  turns recurring quoting and escaping failures into 56 exact-state tasks from 14
  operation families. Controlled payload variants and audited validators isolate
  literal preservation from planning and recovery.
  \item \textbf{A crossed design for mechanism identification.} Generation contract
  and execution transport are varied independently, and fixed-reply replay decomposes
  matched scores into transport damage and contract-conditioned compensation. This
  reveals when aggregate success masks large opposing effects along the command path.
  \item \textbf{Robustness and a measured, not novel, fix.} The transport loss
  persists across effort settings, repeated draws, userlands (GNU versus BSD
  coreutils environments), and held-out payloads. Two obvious fixes, correct
  escaping and a temporary script, each remove the effect entirely; precisely
  because the fixes are trivial, the contribution is the measurement, not the
  repair. Both fixes require the caller to control the boundary, yet our harness
  survey (Table~\ref{tab:harness-survey}) records boundaries applied downstream of
  the stated contract. A matched score alone cannot tell an evaluator whether a fix is
  needed. Typed operations are an exploratory alternative.
\end{enumerate}

These results motivate path-matched model and effort selection and require system
builders to report both the generation contract and execution transport.

\section{Related Work}

\paragraph{Agent and terminal benchmarks.}
General agent benchmarks evaluate web navigation, desktop control, coding, and
interactive execution in realistic environments
\citep{agentbench2024,webarena2023,osworld2024,intercode2023,swebench2024}.
Terminal-focused suites extend this line to command-line workflows and environment
setup \citep{terminalbench2026,envbench2025,terminalworld2026}. Their realism supports end-to-end evaluation, while the contribution of the command
interface remains unresolved. QuoteBench isolates that attribution question.

\paragraph{Shell-command generation and robustness.}
NL2Bash and NLC2CMD formulate natural-language-to-command translation as semantic
parsing or competition-style command generation
\citep{nl2bash2018,nlc2cmd2021}. NL2SH-ALFA adds manually verified data and
execution-based functional-equivalence checks \citep{nl2shalfa2025}. Concurrent
work introduces BashBench, a 952-task benchmark of syntax, functionality, and
robustness for generated Bash programs \citep{bashcoder2026}. Static shell analysis
and long-standing guidance on hostile filenames document the underlying hazards
\citep{shellcheck,wheeler2010filenames}; run on our replies, ShellCheck flags
only 34.6\% of the nested-only failures (versus 11.4\% of the replies that
survive nesting) and misses two-thirds, because each command is individually
well-formed and the fault is in the downstream interpolation. These works score the generated program. QuoteBench fixes that program and changes
the execution transport, separating an incorrect command from a correct command
encoded for the wrong channel.

\paragraph{Action representations and boundaries.}
SWE-agent shows that the agent--computer interface can change coding performance,
and OctoBench separates task completion from compliance with scaffold constraints
\citep{sweagent2024,ding-etal-2026-octobench}. Action Boundary Blindness likewise
shows that conventional success can hide errors in action granularity, scope, and
completion \citep{wang-etal-2026-action}. Tool-use benchmarks emphasize tool
selection and argument construction, while CodeAct and CODESTRUCT change the action
language itself \citep{gorilla2023,toolllm2023,codeact2024,
kim-etal-2026-codestruct}. QuoteBench makes the model-facing contract and downstream transport explicit and
crosses them experimentally. A valid structured call guarantees the envelope, but
shell correctness still depends on the bytes delivered to the executor.

\paragraph{Evaluation validity and deployment safeguards.}
Repeated-sampling studies distinguish one successful trajectory from reliable
repeated execution \citep{taubench2024,monkeys2024}. UTBoost shows that permissive validators can accept incorrect coding-agent patches
\citep{yu-etal-2025-utboost}. QuoteBench therefore audits every validator with
initial states, oracles, naive probes, and targeted mutations. CARE studies
shell-specific pre-execution verification, a complementary safeguard at the command
dispatch boundary \citep{care2026}. A concurrent practitioner report further documents
that shell escaping can reverse the cost and reliability tradeoff between flag-based
and JSON-based CLIs \citep{mastykarz2026cli}. Closest in prescription, concurrent work argues that harness variance can exceed
model variance and that leaderboards should disclose the harness, reporting rank
reversals when the whole scaffold is swapped \citep{harnessdisclosure2026}. Because
it replaces the harness wholesale, including context handling, retry, and
verification, it measures variance but cannot attribute a reversal to a mechanism;
QuoteBench fixes the model output and changes a single parser, so it decomposes the
matched score into transport damage and contract-conditioned compensation. Related
validity audits target reward hacking and protocol gaming rather than the execution
channel \citep{protocolvalidity2026}. Input-side prompt-format sensitivity establishes that scores move with a nuisance
channel variable \citep{sclar2024formatspread}, but because each format is
regenerated, the observed spread cannot be separated into what the channel
destroyed and what the model produced differently. Fixed-output replay is what
makes that separation possible. Together, these results
motivate measuring the command channel as part of the executed system.

\section{The QuoteBench Benchmark}
\label{sec:benchmark}

\subsection{Tasks, scope, and validity}

QuoteBench contains 56 one-shot Bash tasks: 14 operation families, each with a
benign control and three hazardous payload variants. The families cover literal
file content, hostile filenames, regular expressions and globbing, heredocs,
argument and environment passing, JSON and Git state, and two local simulations of
a second shell parser. The hazardous tiers hold the operation fixed while adding
quotes, expansion characters, multiline data, leading dashes, or parser-boundary
conflicts. The same frozen core is used in every campaign reported here.

Each task provides a fixture, an instruction, and a final-state validator. Fixtures
are created without invoking a shell. A model returns one Bash program, which runs in
a fresh working directory with a trimmed environment and a 15-second timeout.
Validators check exact file bytes, received argument vectors, parsed JSON, directory
state, or Git history. They score only the resulting state, so any semantically
correct implementation receives credit. Exit codes cannot substitute: across the
failing executions, 23.4--47.0\% exit zero while leaving the wrong final state
(Appendix Table~\ref{tab:failure-taxonomy}), so a benchmark that trusts return
codes would silently miss up to half of these failures.

\begin{table}[t]
\centering
\small
\caption{One worked QuoteBench task (write-file, tier 3). The
$\langle\!\langle\cdot\rangle\!\rangle$ markers delimit the exact literal
payload and are not part of it. The $\hookleftarrow$ symbol marks a newline that
the payload itself contains. The validator compares exact final bytes, so any
command that reaches the target state passes.
Appendix Table~\ref{tab:coverage-map} maps the full family set to the covered
mechanisms. The artifact contains all 56 instructions, fixtures, oracles, and
validators.}
\label{tab:task-examples}
\begin{tabular}{p{0.16\linewidth}p{0.78\linewidth}}
\toprule
Instruction & Create a file named \texttt{out.txt} whose content is exactly the
text $\langle\!\langle$\texttt{back\textbackslash slash and \`{}date\`{} here}%
$\,\hookleftarrow$\newline\texttt{'single' "double" \$(pwd) 100\%}$\rangle\!\rangle$ followed by a
single trailing newline.\\
\midrule
Hazards & backslash, backtick, command substitution, multiline, mixed quotes\\
Naive command & \texttt{echo "back\textbackslash slash and \`{}date\`{} here
\ldots" > out.txt}; the backticks execute \texttt{date}, \texttt{\$(pwd)}
expands locally, and the payload's inner double quote terminates the string
early\\
Final-state gate & exact bytes of \texttt{out.txt}, including one trailing
newline, with no collateral files\\
One passing command & a single-quoted \texttt{printf} with an embedded-quote
splice (the machine oracle). Any other correct encoding receives equal credit\\
\bottomrule
\end{tabular}
\end{table}

The task families were selected from a pre-release mechanism survey of 86
de-identified incidents in author-owned agent sessions and 412 screened public
reports. This evidence supports coverage of repeatedly observed command-construction
mechanisms. Prevalence estimation and complete shell coverage require a different
study design. QuoteBench focuses on POSIX/Bash command construction. PowerShell, Windows CMD,
authentication, network failures, interactive terminal state, and multi-turn recovery
remain outside the benchmark. Appendix~\ref{app:coverage}
reports the survey counts, inclusion rules, and mechanism-to-family mapping.

We validate both task solvability and scoring strictness. A machine-constructed
oracle solves every task with one command. Benign naive probes pass, whereas their
hazardous counterparts fail on the raw path. We then mutate oracle-produced states by
deleting or altering required artifacts, adding collateral files, restoring files
that should be removed, or changing Git-only state. The validators accept every
oracle and benign probe, and reject every untouched fixture, hazardous probe, and all
197 applicable mutations. As a solvability control, three configurations pass all
56 tasks under the nested transport (Table~\ref{tab:scorecard}), so every task has
a feasible nested solution and the nested arm is not degenerate. These checks cover the specified invalid states. Other validator blind spots may
remain.

\subsection{Contracts and transports}
\label{sec:protocols}

\begin{figure}[t]
\centering
\includegraphics[width=\linewidth]{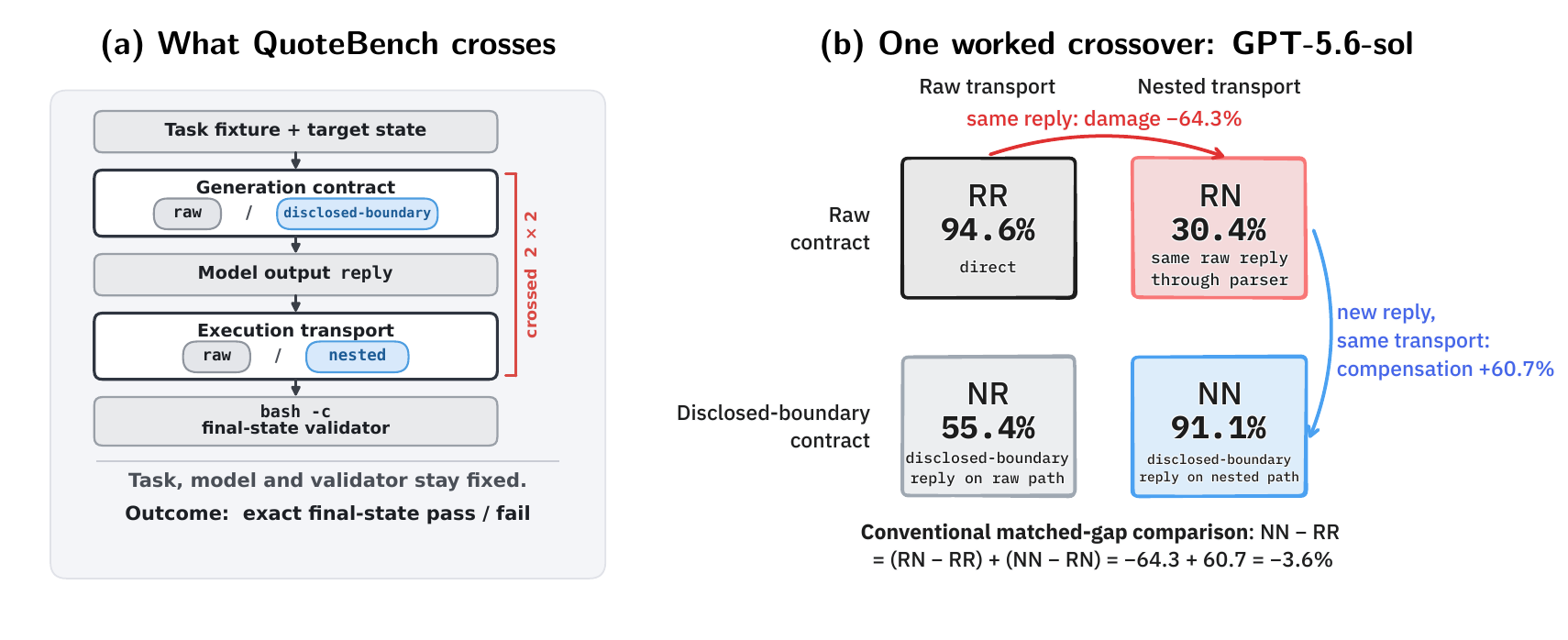}
\caption{\textbf{Crossing generation contract with execution transport.}
Task, model configuration, and validator are fixed across cells. The reply is fixed
within each transport replay pair. Cell labels are contract-then-transport. On the
generation axis, $R$ denotes the raw contract and $N$ the disclosed-boundary
contract. On the transport axis, $R$ denotes raw execution and $N$ the nested
transport. The $RR$ and $NN$ cells are matched. The $RN$ cell measures fixed-reply
damage, and $NN-RN$ is the realized contract-conditioned contrast.
Panel~(b) works the decomposition for GPT-5.6-sol.}
\label{fig:decomposition}
\end{figure}

Figure~\ref{fig:decomposition} previews the crossed design before the result notation:
contract selects the stored reply, transport selects how that reply reaches Bash, and
final-state validation scores the resulting state.

QuoteBench separates the \emph{generation contract}, which tells the model how to
express an action, from the \emph{execution transport}, which determines how that
action reaches a shell. It evaluates two observed model-facing contracts and adds
one controlled transport intervention:

\begin{itemize}
  \item \textbf{Raw contract:} the model emits one Bash program, executed
  verbatim as the script argument to \texttt{bash -c}.
  \item \textbf{Native contract:} the model fills a provider shell-tool call.
  QuoteBench extracts its required \texttt{command} field and executes that string
  on the same raw path, isolating the model-facing representation.
  \item \textbf{Disclosed-boundary contract:} the model is told that its reply
  \texttt{R} will be interpolated into \texttt{bash -c "R"}. The corresponding
  \textbf{nested transport} then adds that parser. Crossing this contract with raw
  and nested transports isolates a boundary that can arise downstream in remote,
  container, or CI commands.
\end{itemize}

A fixed-commit survey of six public agent systems finds both raw and native
model-facing contracts and several downstream transports
(Appendix Table~\ref{tab:harness-survey}). We use \emph{nested} as the name of a
controlled stress condition: it adds the double-quoted parser boundary found in
remote, container, and CI command paths. This boundary is not merely synthetic:
replaying each stored raw reply through a real \texttt{ssh localhost "R"} remote
command reproduces the nested damage to the decimal for seven of eight
configurations and within one task for the eighth
(Appendix Table~\ref{tab:ssh-grounding}), so the perturbation stands in for a
deployment path a model-authored remote wrapper actually produces. Replaying the disclosed-boundary replies through the same real \texttt{ssh}
path completes the 2$\times$2. Real-\texttt{ssh} compensation matches synthetic
nested compensation exactly for five of six replayed configurations;
Gemini-3.1-Flash-Lite differs by one task ($-1.8$ versus $-5.4$ points; Appendix
Table~\ref{tab:ssh-full}). The two non-adapting configurations show no positive
compensation on either path. The raw and
disclosed-boundary generation contracts differ by one sentence that
states this boundary and gives no quoting advice. Appendix~\ref{app:contract-prompts}
gives the prompts verbatim.

All reported primary executions use a pinned, network-disabled GNU/Linux container.
Appendix~\ref{app:userland} repeats the analysis in a BSD/macOS userland. Commands
run only in fresh fixtures with timeouts and collateral-file checks.

\section{Results}
\label{sec:results}

The mechanism analysis rests on eight same-window configurations collected under
one randomized schedule with the effort field omitted (Study A). Broader effort
ladders, a native-tool campaign replayed in both userlands (Study B), and two
private-payload replays extend coverage (Appendix~\ref{app:model-config}).
Table~\ref{tab:scorecard} collapses the ladders to one best-observed row per base
model; model names are provider-public identifiers such as \texttt{gpt-5.6-sol}
(Appendix Table~\ref{tab:study-a-config}).

\subsection{Matched success varies across command paths}
\label{sec:protocol-sets-score}

At their best observed settings, defined as a within-model maximum over single-trial
rungs, three models pass all 56 tasks and
the remaining scores range from 14.3\% to 98.2\% (Table~\ref{tab:scorecard}). Complete measured ladders, including lower operating points and Qwen think toggles,
appear in Figure~\ref{fig:effort-success} and Appendix~\ref{app:model-config}.

The native campaign provides a separate comparison for six evaluated
provider-hosted models. Their
provider-native shell-tool success ranges from 85.7\% to 98.0\%, compared with
95.4--99.3\% on the raw path (Table~\ref{tab:native-absolute}). These values average over every reported effort rung and three trials, whereas
Table~\ref{tab:scorecard} selects one trial from the best observed setting. Native-tool performance is
substantially closer to raw execution than performance under the controlled nested
boundary, although the native effect remains model dependent.

Table~\ref{tab:scorecard} partitions matched-nested success into 14 benign
Control tasks and 42 Hostile tasks. The Hostile LOFO column is a family-jackknife
range (minimum--maximum hostile success over the 14 leave-one-family-out slices),
not a confidence interval. Raw generation itself is close
to saturated at the frontier: the six frontier configurations pass 91.1--100\%
of tasks on the direct path, so raw scores carry almost no discriminative signal.
The entire signal lives on the nested side, which is also the precondition for
masking. What still separates models is how they handle the command path. Matched
nested scores range from 14.3 to 91.1 at the fixed configurations, and realized
compensation ranges from $-5.4$ to $+60.7$; the next two subsections isolate these
effects. The complete
ladders remain visible in Figure~\ref{fig:effort-success}, and all mechanism
estimates use the fixed same-window configurations in Table~\ref{tab:crossover}.

\begin{table}[t]
\centering
\small
\caption{Best-observed QuoteBench scorecard. Each base model contributes the measured setting with the highest matched-nested
All-56 score. Ties prefer \texttt{default}, then lower mean provider-reported output
tokens. \texttt{default}
means that the request omitted the effort field. The number and names of measured
settings differ by provider and appear in Figure~\ref{fig:effort-success} and
Appendix~\ref{app:model-config}. This table is descriptive: each cell is one
stored trial-0 generation and the selection is a within-model maximum over rungs;
Appendix~\ref{app:repeat-generation} reports draw-to-draw spread. The Qwen rows
expose only a think toggle rather than an effort ladder, so their best-observed
setting is taken from the same-window sweep; for Qwen3.5-27B, the only Qwen size
with a ladder row in Table~\ref{tab:v2-ladders}, the resulting one-task difference
is serving-window drift. Fixed same-window configurations support the mechanism
analysis in Table~\ref{tab:crossover}.}
\label{tab:scorecard}
\begin{tabular*}{\textwidth}{@{\extracolsep{\fill}}llrrrr@{}}
\toprule
Model & Best observed setting & Control & Hostile & All 56 (\%) & Hostile LOFO range (\%)\\
\midrule
GPT-5.5 & xhigh & 14/14 & \textbf{42/42} & \textbf{100.0} & [100.0, 100.0]\\
Opus-5 & xhigh & 14/14 & \textbf{42/42} & \textbf{100.0} & [100.0, 100.0]\\
Fable-5 & max & 14/14 & \textbf{42/42} & \textbf{100.0} & [100.0, 100.0]\\
GPT-5.6-sol & high & 14/14 & 41/42 & 98.2 & [97.4, 100.0]\\
Opus-4.8 & max & 12/14 & 41/42 & 94.6 & [97.4, 100.0]\\
Gemini-3.1-Pro & low & 14/14 & 37/42 & 91.1 & [87.2, 89.7]\\
Sonnet-4.6 & high & 9/14 & 27/42 & 64.3 & [61.5, 69.2]\\
Gemini-3.5-Flash & medium & 10/14 & 26/42 & 64.3 & [59.0, 66.7]\\
Haiku-4.5 & medium & 8/14 & 13/42 & 37.5 & [28.2, 33.3]\\
Qwen3.5-27B & non-think & 8/14 & 9/42 & 30.4 & [15.4, 23.1]\\
Qwen3.5-4B & think & 5/14 & 7/42 & 21.4 & [10.3, 17.9]\\
Qwen3.5-9B & non-think & 5/14 & 5/42 & 17.9 & [5.1, 12.8]\\
Gemini-3.1-Flash-Lite & default & 6/14 & 2/42 & 14.3 & [2.6, 5.1]\\
\bottomrule
\end{tabular*}
\end{table}

Measurements use provider-hosted model snapshots identified in
Appendix~\ref{app:model-config}; the same-window mechanism sweep was queried on
2026-07-31 and the effort ladders earlier in July 2026. Hosted deployments may
change under the same identifier, so the query date is part of the result. The
frozen 56-task core is versioned as \texttt{core-v1}.

Controls help separate basic operation competence from literal preservation. At its
best-observed medium setting, Gemini-3.5-Flash passes 10/14 controls and 26/42
hostile tasks. At the lower end, Gemini-3.1-Flash-Lite passes only two of 42 hostile
payloads. Rows are ordered by best-observed matched-nested score. The three perfect rows remain perfect on every hostile leave-one-family-out slice.
GPT-5.6-sol and Opus-4.8 each miss one hostile task at their selected settings. Lower-scoring models
retain distinct family profiles rather than a single shared failure order.

\begin{figure}[t]
\centering
\includegraphics[width=\linewidth]{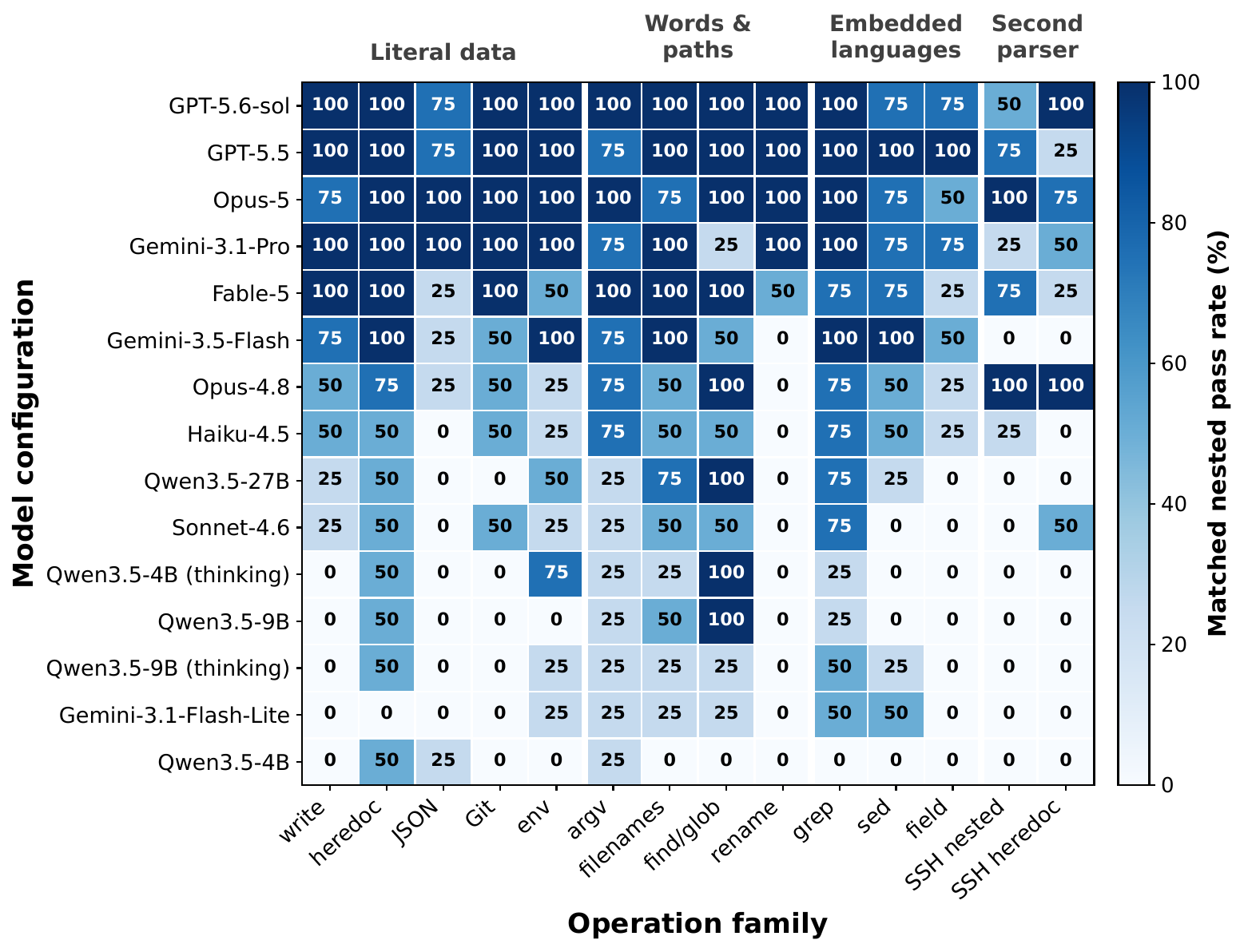}
\caption{\textbf{Matched-nested success varies by operation family across the
frozen configuration sweep.} Each cell is the percentage of a family's four tasks
passed by one stored generation. Rows retain the 15 frozen configurations used for broad coverage; Qwen rows
marked (thinking) enable the think toggle and unmarked Qwen rows are non-thinking,
and Appendix Table~\ref{tab:study-a-config} lists the
queried effort settings and request parameters. Table~\ref{tab:scorecard} separately
summarizes the best observed setting per base model. Columns follow the
mechanism groups in Appendix Table~\ref{tab:coverage-map}. SSH-like tasks use the
local two-shell simulation. Rows are ordered by matched-nested score in this
frozen sweep, which differs from the best-observed order in
Table~\ref{tab:scorecard}.}
\label{fig:family-profile}
\end{figure}

Aggregate rank hides distinct failure profiles. Figure~\ref{fig:family-profile}
shows that models with similar totals fail on different operation families, while
some lower-scoring configurations retain isolated strengths. The heatmap presents
the benchmark at the level users encounter in practice: concrete command families.

For each model, Table~\ref{tab:native-absolute} pools Study B's measured effort
rungs and three trials per cell. The Attempts column is the per-arm denominator.

\begin{table}[H]
\centering
\small
\caption{Absolute GNU replay pass rates in the separate Study-B campaign,
pooled over each model's measured effort ladder and three trials per cell.}
\label{tab:native-absolute}
\begin{tabular}{lrrrr}
\toprule
Model & Attempts & Raw (\%) & Native (\%) & $\Delta$ (pp)\\
\midrule
Opus-4.8 & 840 & 95.4 & 98.0 & $+2.6$\\
Opus-5 & 840 & 98.2 & 97.4 & $-0.8$\\
Fable-5 & 840 & 99.3 & 97.1 & $-2.1$\\
Gemini-3.1-Pro & 504 & 98.8 & 95.0 & $-3.8$\\
GPT-5.6-sol & 672 & 96.9 & 94.3 & $-2.5$\\
Gemini-3.5-Flash & 672 & 95.7 & 85.7 & $-10.0$\\
\bottomrule
\end{tabular}
\end{table}

Across pooled effort rungs, the native-minus-raw change ranges from
$+2.6$ to $-10.0$ points and is smaller than the controlled nested loss for every
model. Appendix Table~\ref{tab:native} reports paired effects, leave-one-family-out
ranges, transitions, BSD comparison, and the failure taxonomy.

\subsection{Transport damage occurs after correct command generation}
\label{sec:crossover}

To identify the failure mechanism, we vary generation contract and execution
transport independently. The task, model configuration, reply, and final-state
validator remain fixed for each replay comparison.

Let $G\in\{R,N\}$ denote the generation contract and $T\in\{R,N\}$ the execution
transport. On the generation axis, $R$ is the raw contract and $N$ is the disclosed-boundary
contract. On the transport axis, $R$ is raw execution and $N$ is the nested transport
(Section~\ref{sec:protocols}). The four cells are $RR$ for a raw reply on raw
transport, $RN$ for a raw reply on nested transport, $NR$ for a disclosed-boundary
reply on raw transport, and $NN$ for a disclosed-boundary reply on nested transport. For one task, $Y_{GT}$ is the corresponding binary final-state
outcome.

\emph{Fixed-reply transport damage} compares $Y_{RN}$ with $Y_{RR}$.
\emph{Contract-conditioned compensation} compares $Y_{NN}$ with $Y_{RN}$. It is
computed from the two stored generations for each task and describes cancellation in
this finite benchmark. The two contrasts sum to the
\emph{matched gap} reported by a conventional matched evaluation:
\begin{equation}
Y_{NN}-Y_{RR}=(Y_{RN}-Y_{RR})+(Y_{NN}-Y_{RN}).
\label{eq:decomp}
\end{equation}
Figure~\ref{fig:decomposition} defines all four cells and works the decomposition
for one configuration. All three quantities are averaged over tasks or operation families
and given in percentage points, written \emph{points} in prose and pp in tables.

The replay reuses stored replies and makes no new model calls. For each of the eight
same-window configurations, the trial-0 reply for every task is executed through both
paths. Effects are averaged over the 14 operation families. Appendix Table~\ref{tab:holm} reports enumerated family-sign sensitivity analyses
with Holm correction. Their scope is the finite, purposively constructed family set.
These intervals and $p$-values quantify variation across the 14 constructed
families, not model-call randomness or a sampled task population.

Moving a fixed raw-generated reply from raw to nested transport costs every
same-window configuration 55.4--73.2 points. The loss is not confined to
adversarial payloads: the 14 benign control tasks alone lose 28.6--57.1 points,
because models emit double-quote-active characters even for ordinary commands. Table~\ref{tab:crossover} gives the
signed effects for those fixed configurations. Its diagonal $RR$ and $NN$ cells are
not the post-selected settings in Table~\ref{tab:scorecard}. All eight effects are
negative, and every
leave-one-family-out estimate remains negative. Reparsing preserves 123 of the 415 direct-path successes, corresponding to
configuration-level retention of 25.0--35.4\%. The remaining 292 become failures,
and the 33 commands that already fail remain failed. Because each pair
reuses the same reply, the added parser accounts for the change in outcome.

The realized generation-by-transport interaction, $(NN-NR)-(RN-RR)$, ranges from
$-7.1$ to $+119.6$ points across the eight configurations
(Table~\ref{tab:crossover}). The wide range shows that the boundary-aware contract changes command behavior in a
transport-specific way. These values characterize the stored generations in this
finite benchmark.

\begin{table}[H]
\centering
\scriptsize
\setlength{\tabcolsep}{3.7pt}
\caption{All four crossover cells for the eight same-window configurations.
Cell notation follows Figure~\ref{fig:decomposition}: generation contract precedes
transport. Cells are pass rates (\%). Effects are percentage points. Damage is
$RN-RR$, compensation is $NN-RN$, and the matched gap is $NN-RR$.
Appendix Table~\ref{tab:effort-crossover} reports the interaction at every measured
rung, and Table~\ref{tab:holm} gives the sensitivity tests.}
\label{tab:crossover}
\begin{tabular*}{\textwidth}{@{\extracolsep{\fill}}lrrrrrrr@{}}
\toprule
Model & $RR$ & $RN$ & $NR$ & $NN$ & Damage & Comp. & Matched gap\\
\midrule
GPT-5.6-sol & 94.6 & 30.4 & 55.4 & 91.1 & $-64.3$ & $+60.7$ & $-3.6$\\
GPT-5.5 & 100.0 & 28.6 & 50.0 & 89.3 & $-71.4$ & $+60.7$ & $-10.7$\\
Opus-5 & 96.4 & 30.4 & 42.9 & 89.3 & $-66.1$ & $+58.9$ & $-7.1$\\
Gemini-3.1-Pro & 98.2 & 25.0 & 33.9 & 80.4 & $-73.2$ & $+55.4$ & $-17.9$\\
Gemini-3.5-Flash & 96.4 & 28.6 & 67.9 & 58.9 & $-67.9$ & $+30.4$ & $-37.5$\\
Opus-4.8 & 91.1 & 26.8 & 62.5 & 57.1 & $-64.3$ & $+30.4$ & $-33.9$\\
Qwen3.5-27B & 85.7 & 30.4 & 83.9 & 30.4 & $-55.4$ & $0.0$ & $-55.4$\\
Gemini-3.1-Flash-Lite & 78.6 & 19.6 & 80.4 & 14.3 & $-58.9$ & $-5.4$ & $-64.3$\\
\bottomrule
\end{tabular*}
\end{table}

At the fixed same-window setting used for this crossover, Gemini-3.5-Flash passes
54/56 tasks in $RR$ but only 8/14 control tasks in $NN$. These counts describe the
fixed mechanism configuration, not the best-observed medium setting in
Table~\ref{tab:scorecard}.

The damage disappears when the boundary is handled correctly. Escaping the reply
at the interpolation point (\texttt{bash -c} \(\langle\)quoted
input\(\rangle\)) reproduces the raw-path outcome exactly for all 448 public
pairs. Replaying the reply as a temporary script does the same for all 448 public
and 126 private-v1 pairs. Neither repair changes raw-path failures: 33 public and
15 private-v1 commands remain failed (Appendix~\ref{app:robustness-detail}).

Matched comparisons can obscure cross-path sensitivity. GPT-5.6-sol's matched
gap is only $-3.6$ points, even though fixed-reply transport loses 64.3 points and
the realized contract-conditioned contrast restores 60.7. The near-zero matched
change is therefore the sum of two large opposing components. The matched $NN$ score
accurately describes its declared path, while off-diagonal replay reveals portability
when that path changes or adds an undisclosed boundary. Appendix~\ref{app:masked-fragility} reports
a descriptive cross-configuration threshold analysis.

The deployment configuration reorders models. The $RR$ and $NN$ orderings agree only
partially: their Kendall rank correlation is $0.57$ (task-cluster bootstrap 95\%
interval $[0.32, 0.82]$, excluding perfect agreement), and $22$ of the $28$
pairwise orderings are stable in at least 95\% of resamples, so the leaderboard is
a bootstrap-supported partial order rather than a fixed ranking. The one reversal
that is unambiguous at this resolution is GPT-5.6-sol versus Gemini-3.5-Flash
(behind by one task under $RR$, ahead by eighteen under $NN$); the count of
reversed pairs is itself uncertain (bootstrap mean $4.6$, 95\% interval $[1, 8]$
of $28$ pairs, Appendix~\ref{app:repeat-generation}).

On a disjoint private set, both models retain negative transport damage and
positive compensation, and three additional draws preserve that sign pattern
(Appendix~\ref{app:robustness-detail}).

\subsection{Matched gains can come from contract-conditioned compensation}
\label{sec:compensation}

The $NN-RN$ contrast fixes the nested transport and compares replies generated under
two contracts that differ by one disclosure sentence. Six of the eight same-window configurations show 30.4 to 60.7 points of realized
compensation, all with family-bootstrap intervals excluding zero (Appendix
Table~\ref{tab:holm}). Qwen3.5-27B shows 0.0 and Gemini-3.1-Flash-Lite $-5.4$. Similar raw scores can accompany substantially different compensation:
Gemini-3.5-Flash and Gemini-3.1-Pro differ by one raw task, yet Pro recovers 25.0
points more. The clause states where the command runs but prescribes no quoting strategy
(Appendix~\ref{app:contract-prompts}).

The compensation is genuine behavioral change, not generic robustness. The same
disclosed-boundary replies that recover the nested path lose 28.6--64.3 points
when replayed on the raw path ($NR$ versus $RR$): the six compensating models
rewrote their commands for the declared boundary and pay for it where the boundary
is absent. The two non-compensating configurations change nothing in either
direction (Qwen3.5-27B $-1.8$, Gemini-3.1-Flash-Lite $+1.8$). Compensation also
concentrates where the hazard is explicit. Payload-quoting families such as
json-write ($+50.0$) and sed-replace ($+46.9$) recover about half their
damage, but implicit hazards remain difficult: find-glob ($-12.5$), grep-count
($+15.6$), and hostile-filenames ($+18.8$) stay broken even under disclosure.

Disclosure, not instruction, carries the effect for capable models. A paired
arm regenerates the advice-free and advice-bearing disclosed contracts in one
serving window, so the contrast has no window confound. At the top of the ladder,
the added escaping advice barely moves matched nested success: GPT-5.6-sol
$-8.9$, GPT-5.5 $+7.1$, and Opus-5 $+3.6$ points. Disclosure alone already elicits
the adaptation. In the middle it makes the largest difference: Sonnet-4.6 gains $+25.0$,
Haiku-4.5 $+12.5$, and Opus-4.8 $+7.1$ points from the advice. At the bottom
neither contract helps (Qwen3.5-27B and Gemini-3.1-Flash-Lite $+1.8$). Boundary
advice thus barely moves the top, changes middle-tier outcomes the most, and does
not move the bottom (Appendix
Table~\ref{tab:advice-arm}).

The adaptation is conditioned on the declared grammar rather than applying a fixed
defense. A crossed arm discloses either a double-quote or a single-quote wrapper
and replays each stored reply through both, forming a 2$\times$2 of disclosed
against executed grammar. Capable models pass far more on the grammar they were
told than on the other: GPT-5.6-sol passes 53/56 of its single-disclosed replies
on the single-quote wrapper but only 10/56 on the double-quote one, and its
diagonal (matched) advantage over the anti-diagonal is $+80.4$ points. The
advantage separates the same top, middle, and bottom groups as the matched-nested
scores: $+80.4$, $+77.7$, and $+65.2$ at the top, $+18.8$ to $+25.0$ in the middle, and $+0.0$ (Qwen3.5-27B) to $-19.6$
(Gemini-3.1-Flash-Lite) at the bottom (Appendix
Table~\ref{tab:grammar-crossover}). This is contract-conditioned behavioral adaptation to the declared grammar, not a
memorized double-quote fix.

\begin{figure}[t]
\centering
\includegraphics[width=\linewidth]{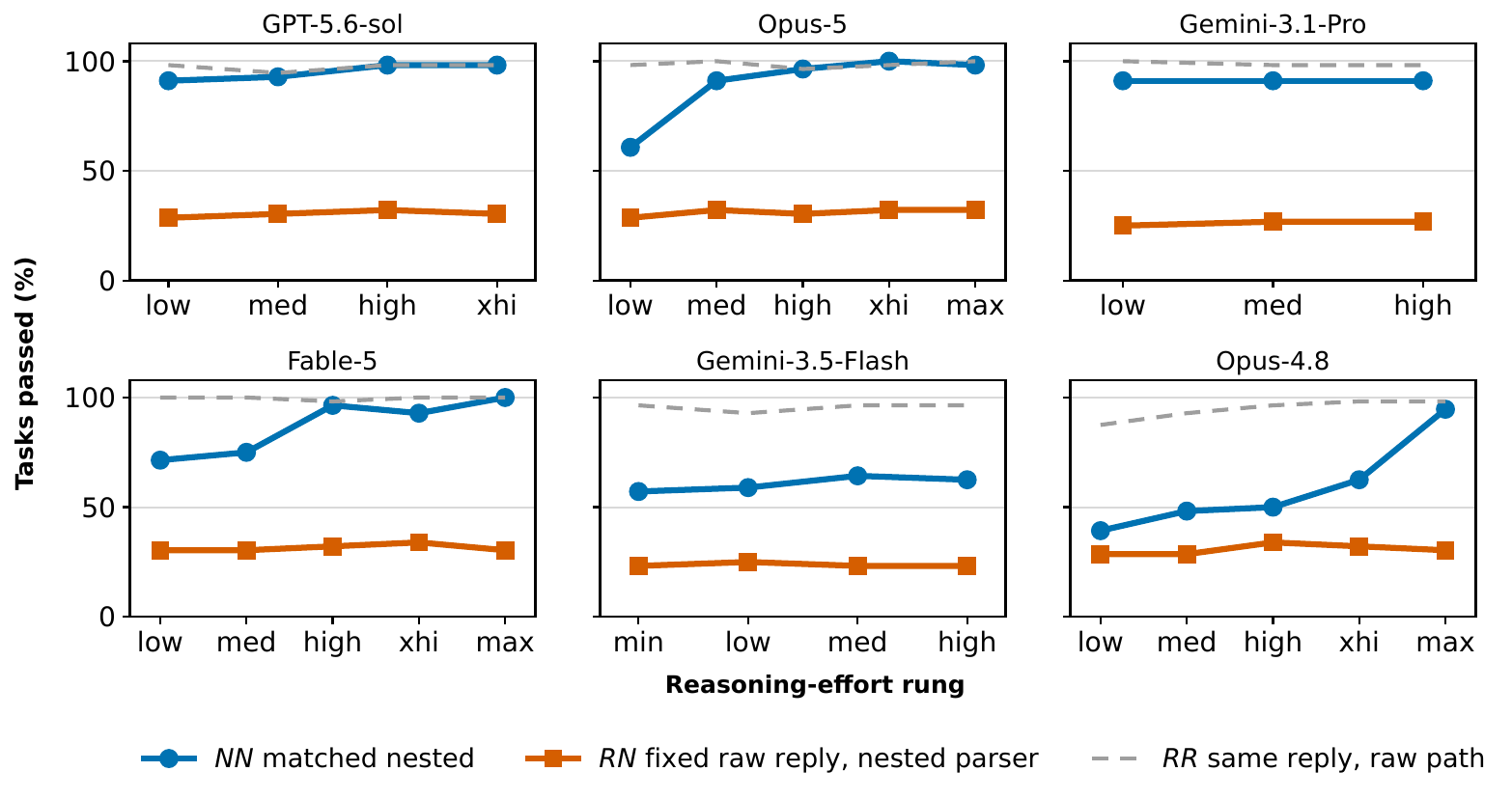}
\caption{\textbf{Where effort raises matched success, the nested-replay pass
rate changes little.} Each point uses 56 tasks, trial 0, and the GNU replay. The $RR$ value is raw success, $RN$ is the nested-replay pass rate of
raw-conditioned replies, and $NN$ is matched nested success. The $NN-RN$ gap is
compensation. Appendix Table~\ref{tab:effort-crossover} reports
every rung, including the byte-identical
Gemini-3.1-Flash-Lite settings.}
\label{fig:effort-crossover}
\end{figure}

Six configurations provide raw and disclosed-boundary replies at every effort rung,
yielding 26 crossover points from stored generations
(Figure~\ref{fig:effort-crossover} and Appendix
Table~\ref{tab:effort-crossover}). The nested-replay pass rate of raw-conditioned
replies stays between 23.2\% and 33.9\%, moves by at most 5.4 points within any one
ladder, and accompanies damage of $-58.9$ to $-75.0$ points. These trajectories
summarize one stored generation at each rung.

Most matched-score movement comes from the contract-conditioned contrast. It rises
from $+10.7$ at low to $+64.3$ at max for Opus-4.8 and from $+32.1$ at low to
$+66.1$ at max for Opus-5. Interior rungs are not monotone
(Figure~\ref{fig:effort-crossover}).
Gemini-3.1-Flash-Lite returns byte-identical replies at all four settings and therefore
contributes a single observed operating point.

Because $RN$ varies little while $NN$ sometimes climbs, the matched gap can narrow
without an improved nested-replay pass rate. The pattern varies by model:
Gemini-3.1-Pro remains at 91.1\% across its three rungs, and Gemini-3.5-Flash moves
only 7.1 points.

Opus-4.8 shows the masking effect when matched success does climb. Its matched gap
moves from $-48.2$ points at low to $-3.6$ at max, while damage grows from $-58.9$ to
$-67.9$. The raw arm also improves by 10.7 points, but the nested-replay pass rate
ends near where it began. A matched evaluation would attribute the improvement to repair, whereas fixed-reply
replay shows that cross-path portability remains essentially unchanged.

\begin{figure}[t]
\centering
\includegraphics[width=\linewidth]{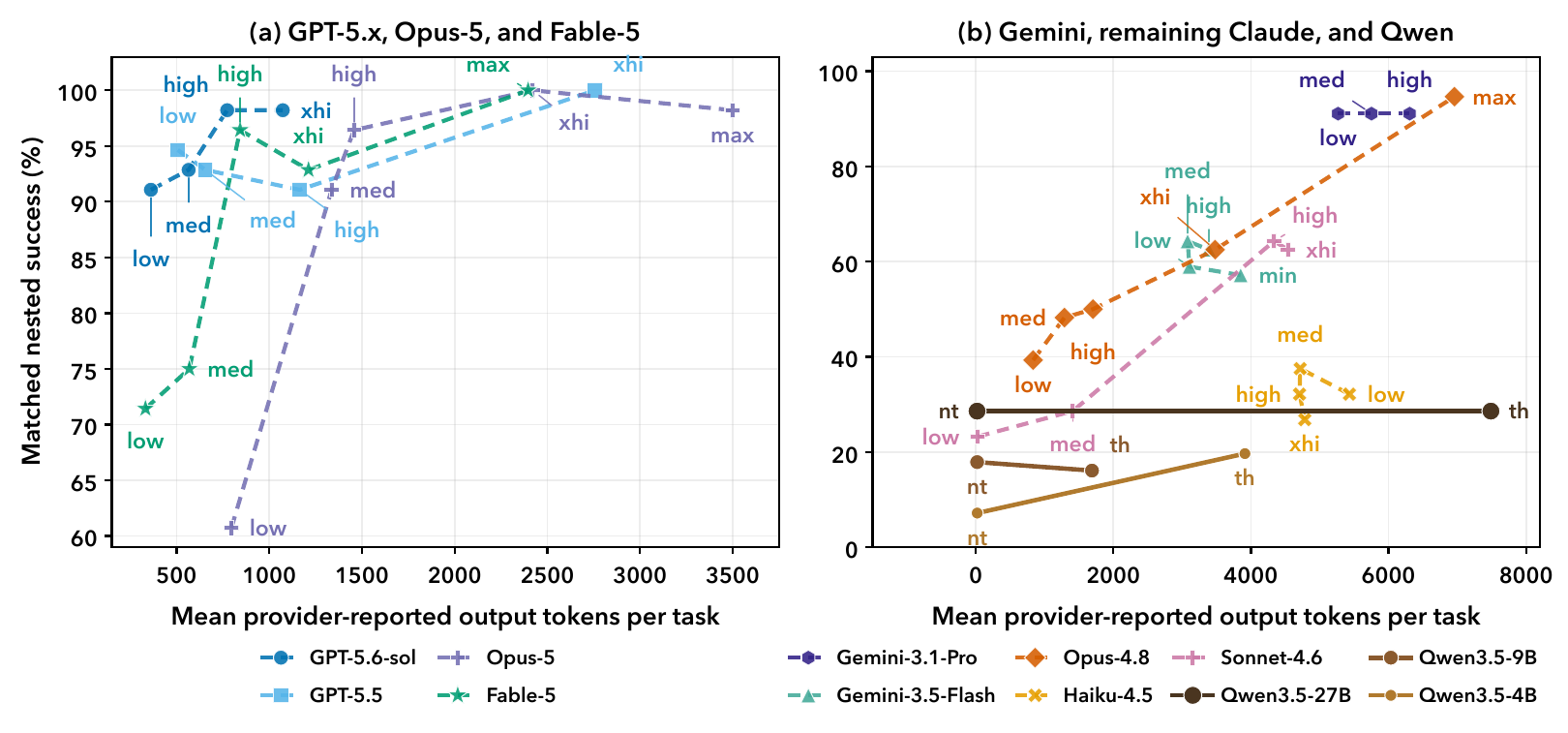}
\caption{\textbf{Effort settings are provider-specific operating points.} Each point
shows trial-0 matched-nested success on the 56-task core against mean
provider-reported output tokens per task, including reported hidden reasoning.
Dashed segments follow the provider's declared order. Backward or dominated segments
show measured non-monotonicity. Panels group model families for legibility, and Qwen
sizes contribute two-point think-toggle trajectories.}
\label{fig:effort-success}
\end{figure}

Each measured effort rung is a deployment-relevant operating point under the frozen
benchmark, not an estimate of effort's causal effect. Figure~\ref{fig:effort-success}
shows that labels map to different token budgets across models and that several
ladders are non-monotonic.

An unset effort field maps to different parts of each provider's ladder. Across the
seven same-window configurations with both measurements, Opus-4.8's unset arm
resembles xhigh, Opus-5's resembles medium, and Gemini-3.1-Pro's falls below its
entire ladder. Users and evaluators should compare operating points through measured
behavior. Appendix~\ref{app:effort-detail}
reports the exact rung values and calibration table.

\subsection{The mechanism transfers across payloads and sampled replies}
\label{sec:robustness}

The mechanism is not a property of the published payloads. The private-v2
crossover repeats the design on 42 unpublished hostile payloads:
raw- and disclosed-boundary-contract calls for GPT-5.6-sol and Opus-4.8 were
interleaved within one serving window, and each stored reply was replayed through
both transports with the public final-state validators.

\begin{table}[t]
\centering
\scriptsize
\caption{Private-v2 crossover on 42 hostile payloads under the single-clause
disclosed-boundary contract. The private set is hostile-only and not
difficulty-matched to the public core, so absolute rates are interpreted within
this set. Cells are pass rates. Effects are percentage points. Damage is
$RN-RR$, compensation is $NN-RN$, interaction is $(NN-NR)-(RN-RR)$, and the
matched gap is $NN-RR$.}
\label{tab:heldout}
\begin{tabular}{lrrrrrrrr}
\toprule
Model & $RR$ & $RN$ & $NR$ & $NN$ & Damage & Comp. & Interaction & Matched gap\\
\midrule
GPT-5.6-sol & 92.9 & 19.0 & 50.0 & 97.6 & $-73.8$ & $+78.6$ & $+121.4$ & $+4.8$\\
Opus-4.8 & 92.9 & 16.7 & 59.5 & 42.9 & $-76.2$ & $+26.2$ & $+59.5$ & $-50.0$\\
\bottomrule
\end{tabular}
\end{table}

Both models pass 92.9\% of the private tasks on the direct raw path, then lose 73.8
and 76.2 points when the same replies cross the added parser. Compensation remains
model dependent: GPT-5.6-sol recovers 78.6 points under the boundary-aware contract,
whereas Opus-4.8 recovers 26.2. Every leave-one-family-out slice preserves negative
transport damage.

Additional sampling preserves the same interpretation. Three more generations for
eight private tasks produce different reply text in 19 of 32 task--contract cells,
yet all four draws retain negative damage and positive compensation for both models.

A separate private-v1 replay tests a practical bypass on 42 tasks and three models,
executing each stored raw-contract reply through the raw path, the nested
wrapper, and a temporary Bash script.

\begin{table}[H]
\centering
\small
\caption{Private-v1 fixed raw replies under three transports on 42 tasks. This campaign predates the private-v2 crossover in Table~\ref{tab:heldout}. The two
tables therefore use different generations. The first three columns report tasks passed out
of 42. Script gain is the temporary-script rate minus the nested-wrapper rate, in
percentage points.}
\label{tab:script-bypass}
\begin{tabular}{lrrrr}
\toprule
Model & Raw & Nested wrapper & Temporary script & Script gain (pp)\\
\midrule
GPT-5.6-sol & 41/42 & 8/42 & 41/42 & $\mathbf{+78.6}$\\
Opus-4.8 & 40/42 & 7/42 & 40/42 & $\mathbf{+78.6}$\\
Qwen3.5-27B & 30/42 & 9/42 & 30/42 & $\mathbf{+50.0}$\\
\bottomrule
\end{tabular}
\end{table}

Executing the replies from temporary scripts reproduces the raw-path outcome for
every model--task pair. It recovers 87 commands that fail only under the wrapper. The 15 commands that fail
directly remain unresolved. Appendix~\ref{app:heldout}
reports the private design and validator checks, and Appendix~\ref{app:robustness-detail}
reports the repeated draws, script-bypass records, and typed-operation pilot.

\section{Discussion}

The results support two reporting practices for command paths that wrap or reparse
model output. First, report the generation contract together with the execution path.
The crossed design shows that the generation contract can change the matched score,
while off-diagonal replay reveals whether replies remain portable across paths. Second,
structured actions remove one quoting boundary while leaving payload-level
representation errors possible. In the typed pilot, eleven of 36 programs fail and
ten leave the wrong final state, most often because the model copies instruction
delimiters into the payload (Appendix~\ref{app:robustness-detail}).

The command interface is part of the evaluated system, not neutral plumbing.
Vendors should report the generation contract, deployed execution path, selected
operating point, and family-level failures. Users should compare models and effort
settings on that path because provider ladders are non-monotonic and defaults map to
different operating points.

Ignoring the path changes which model wins: selecting by raw success picks
GPT-5.5 (56/56 raw), which reaches 50/56 on the nested path, whereas the
path-aware pick reaches 51/56. The regret is small at the saturated frontier, but the
reversed top rank. The scorecard and crossover answer different questions. Table~\ref{tab:scorecard}
supports operating-point selection, while Table~\ref{tab:crossover} diagnoses path
sensitivity for fixed configurations. Neither is a controlled compute ranking.
Deployment reports should publish both the selected point and the measured ladder.

Userland changes are smaller than the added-parser effect, but they are not always
negligible. In the current Study-A campaign, fixed-reply damage remains negative in both
BSD/macOS and GNU/Linux. Corresponding crossover cells differ by at most 3.6 points
for six evaluated provider-hosted configurations and by 7.1--12.5 points for the
other two.

An earlier frozen BSD-live campaign reveals model-specific dialect affinity. The identical stored commands improve on GNU for both Qwen3.5-27B settings and
Gemini-3.1-Pro, while Fable-5 and Gemini-3.5-Flash retain higher success on BSD.
Opus-4.8 is higher on BSD in the raw arm and tied in the nested arm. Shifts reach 8.9
points and partially reorder the models under both generation contracts. Because all commands
were elicited in BSD/macOS sessions, this analysis measures cross-userland transfer,
not what a model would generate when explicitly targeting GNU. Study B hints at a
contract-by-userland interaction. Gemini-3.1-Pro and Fable-5 flip from positive
native-minus-raw changes on BSD to negative on GNU, though the shift does not
survive the sensitivity analysis (Appendix~\ref{app:family-detail}). Command benchmarks should
therefore report their userland and replay stored commands across the environments
they claim to support (Appendix~\ref{app:userland}).

The decomposition should extend beyond shell, though we test only one boundary
here. Any pipeline that transforms generated output before execution defines the
same four cells. A JSON tool-call boundary is a second instance. Replaying each stored raw reply
through a naive JSON string embedding causes losses from 51.8 to 66.1 points. The serializer
re-parses the same double quotes and backslashes as the shell, while a correct
round-trip serializer costs exactly zero (Appendix~\ref{app:json-boundary}). The
mechanism is an unescaped transform, not any transform, and the decomposition
transfers to a non-shell boundary.

Where a deployed path adds an interpolating shell layer, the first-line fix is
harness-side: escaping the reply at the interpolation point restores every
raw-path success in our replays. Where the boundary is not under the caller's
control, as in remote or CI patterns, a temporary script preserves the program
boundary at the cost of a file lifecycle, and boundary disclosure lets capable
models compensate (Appendix~\ref{app:robustness-detail}).

\section{Conclusion}

QuoteBench shows that matched execution scores can hide post-generation failure.
Replaying fixed replies through one added parser lowers success by 55.4--73.2 points
across all same-window configurations, while contract-conditioned generation recovers
30.4--60.7 points for six configurations. At best observed settings, three models
score 56/56 and others span 14.3\%--98.2\%. Trial-0 ladders shift nested-replay
pass rates by at most 5.4 points.

\section*{Limitations}

QuoteBench isolates one mechanism: one-shot Bash generation under quotation and
interpolation hazards. Its 14 constructed families support mechanism attribution,
and the nested transport reproduces a real \texttt{ssh} remote-execution
boundary (Table~\ref{tab:ssh-grounding}) rather than a claim about how often such
boundaries occur; results characterize this benchmark and stored-reply
portability, not deployment prevalence. Causal claims rest on fixed replies in
the eight same-window configurations, while effort-ladder rungs rely on a single
stored generation per task and effort labels are not comparable compute budgets.
The native-tool campaign is observational. Held-out payloads test transfer to
unseen literals without difficulty matching, the typed-operation study is limited
to six naturally typeable families, and other shells and multi-turn recovery
remain open.

\section*{Broader Impact Statement}

QuoteBench executes untrusted model output, so the released harness runs each
attempt in a fresh fixture inside a timeout-bounded, network-disabled container,
and incident evidence is released only as de-identified mechanism
classifications. Publishing the frozen core creates a contamination risk; we
treat it as a versioned audit set and hold out regenerated private variants. Each
released task file also embeds a fixed canary GUID, recorded in the repository, so
downstream contamination checks have a known token to search for. The
benchmark introduces no shell capability beyond routine coding-agent operations.

\section*{Acknowledgements}
We gratefully acknowledge Jiafu Tang and Ziyu Zhou for providing access to the
Gemini and GPT APIs, respectively. Their support enabled the experiments reported
in this work.

\bibliographystyle{tmlr}
\bibliography{references}


\appendix
\section{Benchmark Construction, Coverage, and Contracts}
\label{app:coverage}

Two mechanism surveys guided the family design. The internal survey contains 86
de-identified incidents from author-owned coding-agent sessions: 50 Codex incidents
and 36 Claude incidents. The public survey screened 412 candidates, read 34 in full,
retained 17 model-level POSIX/Bash command-construction incidents, classified 10
harness failures separately, and excluded seven cases outside scope. Every retained
model-level incident maps to a mechanism represented in the 14-family core. The surveys document mechanism coverage. Prevalence and complete shell coverage remain outside their purpose. Representative retained reports
span Claude Code, Codex, Gemini CLI, Warp, and OpenHands
\citep{claudecodeheredoc2026,claudecodegh2026,codexcontract2026,
geminiescaping2025,warpheredoc2025,openhandsheredoc2024}.

The released survey record includes the tracker query, inclusion decision, and
mechanism code for every candidate. We retain a report as model-level when the model
constructs the POSIX command, the failure concerns literal or argument semantics,
execution permits final-state scoring, and command-level evidence identifies the
mechanism.
Product-side rewrites are classified separately as harness regressions. Five of the
seventeen retained incidents target a second parser, such as an SSH remote or inner
\texttt{shell -c}, matching the nested boundary of \S\ref{sec:protocols}.

\begin{table}[H]
\centering
\scriptsize
\caption{Mechanism groups used to construct the frozen core; the mapping documents coverage of the surveyed failure mechanisms. Benchmark analyses weight the 14 operation families equally.}
\label{tab:coverage-map}
\begin{tabular}{p{0.25\linewidth}p{0.30\linewidth}p{0.35\linewidth}}
\toprule
Mechanism group & Representative failures & QuoteBench families\\
\midrule
Literal quote and expansion & Apostrophes, double quotes, dollars, backticks,
multiline payloads & write-file, JSON writing, Git commit, environment passing,
heredoc writing\\
Word splitting and path semantics & Spaces, globs, leading dashes, hostile filenames,
argument boundaries & argv passing, hostile filenames, find/glob, bulk rename\\
Embedded-language escaping & Regex versus literal matching, sed replacement, AWK
string processing & grep count, sed replace, field lookup, JSON writing\\
Second parser or remote-like expansion & Local expansion before a second shell,
argument joining, heredoc transport & SSH-like nested execution, SSH-like heredoc\\
Command-boundary representation & Command string, shell stdin, temporary file, argv,
provider tool schema & raw/nested crossover, native-tool study, script bypass, typed
pilot\\
\bottomrule
\end{tabular}
\end{table}

\subsection{Surveyed command boundaries}

The main text uses only the distinction needed for the intervention. Table
\ref{tab:harness-survey} records the implementation evidence behind that
classification. Contract denotes what the system asks the model to produce. Observed boundary denotes what the harness subsequently does with the reply.

\begin{table}[H]
\centering
\scriptsize
\caption{Command boundaries in six public agent systems, inspected at fixed
commits. \emph{Contract} is what the system asks the model to produce. A nested boundary arises on a separate axis from what the command targets. \emph{Observed boundary} is what the harness then does with the reply. The
classification concerns the parser boundary only. Sources:
\citet{codexsource2026,sweagentsource2026,langchainshellsource2026,
terminalbenchsource2026,openhandsargvsource2026,autogendockersource2026}.}
\label{tab:harness-survey}
\begin{tabular}{p{0.115\linewidth}p{0.065\linewidth}p{0.395\linewidth}p{0.315\linewidth}}
\toprule
System & Contract & Source anchor & Observed boundary\\
\midrule
Codex & native & \texttt{core/src/shell.rs:20--30}, commit \texttt{fa1d4c4} & command string becomes \texttt{shell -c/-lc R}\\
SWE-agent & raw & \texttt{agents.py:936--967} and \texttt{swe\_env.py:197--222}, commit \texttt{3ea751c} & agent action enters a persistent Bash session\\
LangChain & native & \texttt{shell\_tool.py:217--232,491--515}, commit \texttt{b3a6d9a} & structured \texttt{command: string} is written to shell stdin\\
Terminal-Bench & raw & \texttt{tmux\_session.py:26--33,75--173}, commit \texttt{d28711d} & command/key strings enter an interactive shell through tmux\\
OpenHands & native & \texttt{acp-command.ts:11--45,65--147}, commit \texttt{850bd64} & human-readable command is tokenized to argv, and spawn has no shell\\
AutoGen & native & \texttt{\_docker\_code\_executor.py:327--363}, commit \texttt{027ecf0} & generated code is written to a temporary file and invoked by argv\\
\bottomrule
\end{tabular}
\end{table}

The nested transport is a synthetic stand-in for a real remote-execution
boundary. Table~\ref{tab:ssh-grounding} confirms it behaves like one: each stored
raw reply, replayed through an actual \texttt{ssh localhost "R"} command in a
loopback-\texttt{sshd} container (a zero-call execution, the frozen runner image
plus a local \texttt{sshd}), loses the same points it loses under the synthetic
nested transport.

\begin{table}[h]
\centering
\small
\caption{Real-\texttt{ssh} grounding. Each stored raw reply is replayed through
\texttt{bash -c} and through a real \texttt{ssh localhost "R"} remote command;
\emph{ssh damage} is the second minus the first. It matches the synthetic nested
damage to the decimal for seven of eight configurations; the disclosed-reply
cells are grounded in Appendix Table~\ref{tab:ssh-full}. Zero model calls.}
\label{tab:ssh-grounding}
\begin{tabular}{lrrrr}
\toprule
Configuration & Raw & \texttt{ssh} & ssh damage & Nested damage\\
\midrule
GPT-5.6-sol & 94.6 & 30.4 & $-64.3$ & $-64.3$\\
GPT-5.5 & 100.0 & 28.6 & $-71.4$ & $-71.4$\\
Opus-5 & 96.4 & 30.4 & $-66.1$ & $-66.1$\\
Gemini-3.1-Pro & 98.2 & 26.8 & $-71.4$ & $-73.2$\\
Gemini-3.5-Flash & 96.4 & 28.6 & $-67.9$ & $-67.9$\\
Opus-4.8 & 91.1 & 26.8 & $-64.3$ & $-64.3$\\
Qwen3.5-27B & 85.7 & 30.4 & $-55.4$ & $-55.4$\\
Gemini-3.1-Flash-Lite & 78.6 & 19.6 & $-58.9$ & $-58.9$\\
\bottomrule
\end{tabular}
\end{table}

The released tasks are generic programmatic distillations carrying no private
commands and no personal data. The public core excludes PowerShell, Windows CMD,
interactive terminal state, real SSH networking and authentication, and complete
multi-turn recovery. Generated variants and future shell tracks are versioned
separately so the 56-task core remains auditable.

\subsection{Generation contracts and command boundaries}
\label{app:contract-prompts}

The task instruction and the common shell rules are byte-identical across all
Study-A source arms. The system prompts differ only in the execution-contract clause. Each contract differs from the next by a single clause, and that clause
states the execution environment rather than explaining how to quote for it.

The raw clause is:

\begin{quote}\small\ttfamily
Reply with exactly one bash command and nothing else -{}-{}- no explanation, no markdown
fences. Your reply is passed verbatim as the single argument to bash -c and
executed in the task's working directory.
\end{quote}

The single-clause disclosed-boundary contract used for every matched nested result
in this paper is:

\begin{quote}\small\ttfamily
Reply with exactly one bash command and nothing else -{}-{}- no explanation, no markdown
fences. Your reply R is not executed directly: it is interpolated inside double
quotes into an outer command, producing the string bash -c "R", and that string is
executed in the task's working directory.
\end{quote}

A character-level diff of the two full system prompts shows that the sentence
beginning ``Your reply'' is the only text that changes. Both arms then continue with the same shared rules. The reply may use pipes,
\texttt{\&\&}, \texttt{;}, or quoted embedded newlines, but it must remain one
\texttt{bash -c} payload. Interactive programs and helper script files are excluded,
and text between the $\langle\!\langle\cdot\rangle\!\rangle$ markers is exact
literal text. Compensation
therefore measures the total effect of stating the parsing environment, with no
format constraint and no escaping instruction attached.

\paragraph{Other evaluated contracts.}
The typed-operation pilot must state the interface semantics, but its request to
preserve every literal character, including trailing newlines, provides mild coaching. The pilot is therefore exploratory. The native contract also mentions a
Bash script payload, a difference included in Study B's declared total-effect
estimand. The marker note and shared rules are common to all arms.

Appendix~\ref{app:reproduction} lists the packaged evidence and includes the exact
contract string literals used by the harness, allowing direct verification against
the source.

\section{Model Configurations and Effort Details}
\label{app:model-config}

Table~\ref{tab:campaign-map} lists every measurement campaign the paper draws on
and which results it feeds.

\begin{table}[h]
\centering
\scriptsize
\caption{Campaign map. All replays are zero-call executions of stored replies in
the pinned container.}
\label{tab:campaign-map}
\begin{tabular}{llll}
\toprule
Campaign & Generations & Design & Feeds\\
\midrule
Study A same-window sweep & 8 configs $\times$ 56 $\times$ 2 contracts & one randomized window, effort unset & Tables~\ref{tab:crossover}, \ref{tab:holm}\\
Study A ladder sweep & 44 rungs, 11 configs & per-provider windows & Tables~\ref{tab:scorecard}, \ref{tab:v2-ladders}\\
Study A rung crossover & 30 rungs, 7 configs, 26 crossover pts & replay both transports & Table~\ref{tab:effort-crossover}, Fig.~\ref{fig:effort-crossover}\\
Public three-draw repeat & 8 configs $\times$ 56 $\times$ 2 $\times$ 2 draws & same design as the sweep & Appendix~\ref{app:repeat-generation}\\
Study B native tool & 8{,}736 generations, 17{,}472 replays & observational, both userlands & Tables~\ref{tab:native-absolute}, \ref{tab:native}, \ref{tab:failure-taxonomy}\\
Private-v2 crossover & 2 models $\times$ 42 hostile payloads & one serving window & Table~\ref{tab:heldout}\\
Private-v1 replay & 3 models $\times$ 42 tasks & earlier generations; script bypass & Table~\ref{tab:script-bypass}\\
BSD-live legacy & 6 configurations & BSD-elicited, GNU-replayed & Table~\ref{tab:dialect-transfer}\\
Real-ssh grounding & 8 configs $\times$ 56 & \texttt{ssh localhost} replay & Table~\ref{tab:ssh-grounding}\\
Advice arm & 8 configs $\times$ 56 $\times$ 2 & same-window paired advice contrast & Table~\ref{tab:advice-arm}\\
Grammar crossover & 8 configs $\times$ 56 $\times$ 2 disclosed & replay-only wrapper 2$\times$2 & Table~\ref{tab:grammar-crossover}\\
Real-ssh full crossover & 6 configs $\times$ 56 $\times$ 2 & disclosed replies on real ssh & Table~\ref{tab:ssh-full}\\
JSON boundary & 6 configs $\times$ 56 & serializer replay & Table~\ref{tab:json-boundary}\\
\bottomrule
\end{tabular}
\end{table}

Mechanism analysis is restricted to the eight same-window configurations. The
broader matched-outcome ladders remain useful for descriptive operating-point
comparison, but no off-diagonal cell is reconstructed across serving windows.
Table~\ref{tab:default-calibration} shows why an omitted effort field cannot be
interpreted as a common neutral rung.

\begin{table}[h]
\centering
\small
\caption{Calibration of the unset-effort sweep arm against each
configuration's labelled ladder. Distance is unset success minus the lowest-rung
success, in percentage points. The two measurements come from different serving
windows, so small differences are descriptive only.}
\label{tab:default-calibration}
\begin{tabular*}{\textwidth}{@{\extracolsep{\fill}}lrll@{}}
\toprule
Model & Unset (\%) & Nearest rung & Distance to lowest rung\\
\midrule
GPT-5.6-sol & 91.1 & low & 0.0\\
GPT-5.5 & 89.3 & high & $-5.4$\\
Opus-5 & 89.3 & medium & $+28.6$\\
Gemini-3.1-Pro & 80.4 & none within the ladder & $-10.7$\\
Gemini-3.5-Flash & 58.9 & low & $+1.8$\\
Opus-4.8 & 57.1 & xhigh & $+17.9$\\
Gemini-3.1-Flash-Lite & 14.3 & all four rungs tie & 0.0\\
\bottomrule
\end{tabular*}
\end{table}

\subsection{Effort ladders and model configurations}
\label{app:effort-detail}

Tables~\ref{tab:v2-ladders} and~\ref{tab:study-a-config} document the measured
operating points. The first reports each model's exposed settings and outcomes; the
second records exact model identifiers and request parameters. Provider labels are
within-model controls, not common compute units. Qwen exposes a think toggle rather
than a multi-rung effort parameter. For Haiku-4.5, the output-token means are strongly
right-skewed; the corresponding medians are 1{,}206, 1{,}560, 936, and 1{,}283 tokens,
so the non-monotonic budget ordering persists under a robust summary.

\begin{table}[H]
\centering
\scriptsize
\caption{Matched-nested effort ladders under the disclosed-boundary contract. Within
each row, success rates and mean provider-reported output tokens follow the setting
order in the second column. Each point uses trial 0 over 56 tasks.}
\label{tab:v2-ladders}
{\setlength{\tabcolsep}{4.2pt}
\begin{tabular*}{\textwidth}{@{\extracolsep{\fill}}lp{0.30\textwidth}p{0.22\textwidth}p{0.23\textwidth}@{}}
\toprule
Model & Settings (in order) & Success (\%) & Mean output tokens\\
\midrule
GPT-5.6-sol & low / medium / high / xhigh & 91.1 / 92.9 / 98.2 / 98.2 & 362 / 565 / 773 / 1{,}073\\
GPT-5.5 & low / medium / high / xhigh & 94.6 / 92.9 / 91.1 / 100.0 & 507 / 655 / 1{,}164 / 2{,}757\\
Opus-5 & low / medium / high / xhigh / max & 60.7 / 91.1 / 96.4 / 100.0 / 98.2 & 796 / 1{,}336 / 1{,}458 / 2{,}421 / 3{,}499\\
Fable-5 & low / medium / high / xhigh / max & 71.4 / 75.0 / 96.4 / 92.9 / 100.0 & 332 / 569 / 843 / 1{,}212 / 2{,}396\\
Opus-4.8 & low / medium / high / xhigh / max & 39.3 / 48.2 / 50.0 / 62.5 / 94.6 & 835 / 1{,}291 / 1{,}706 / 3{,}481 / 6{,}960\\
Gemini-3.1-Pro & low / medium / high & 91.1 / 91.1 / 91.1 & 5{,}267 / 5{,}753 / 6{,}308\\
Sonnet-4.6 & low / medium / high / xhigh & 23.2 / 28.6 / 64.3 / 62.5 & 28 / 1{,}411 / 4{,}337 / 4{,}539\\
Gemini-3.5-Flash & minimal / low / medium / high & 57.1 / 58.9 / 64.3 / 62.5 & 3{,}851 / 3{,}108 / 3{,}081 / 3{,}389\\
Haiku-4.5 & low / medium / high / xhigh & 32.1 / 37.5 / 32.1 / 26.8 & 5{,}432 / 4{,}717 / 4{,}708 / 4{,}783\\
Qwen3.5-27B & non-thinking / thinking & 28.6 / 28.6 & 20 / 7{,}489\\
Gemini-3.1-Flash-Lite & minimal / low / medium / high & 14.3 / 14.3 / 14.3 / 14.3 & 20 / 20 / 20 / 20\\
\bottomrule
\end{tabular*}}
\end{table}

\begin{table}[H]
\centering
\scriptsize
\caption{Study-A model identifiers and request parameters. The effort column lists
exactly the settings queried; sweep arms omitted the effort field. Temperature is
reported only where the interface accepts it.}
\label{tab:study-a-config}
{\setlength{\tabcolsep}{4.0pt}
\begin{tabular*}{\textwidth}{@{\extracolsep{\fill}}p{0.145\linewidth}p{0.235\linewidth}p{0.235\linewidth}p{0.295\linewidth}@{}}
\toprule
Display name & Model identifier & Effort settings queried & Decoding parameters\\
\midrule
GPT-5.6-sol & \texttt{gpt-5.6-sol} & low, medium, high, xhigh &
max output tokens 16{,}000, temperature not sent\\
GPT-5.5 & \texttt{gpt-5.5} & low, medium, high, xhigh &
max output tokens 16{,}000, temperature not sent\\
Opus-5 & \texttt{claude-opus-5} & low, medium, high, xhigh, max &
provider defaults, no sampling or length control sent\\
Opus-4.8 & \texttt{claude-opus-4-8} & low, medium, high, xhigh, max &
provider defaults, no sampling or length control sent\\
Fable-5 & \texttt{claude-fable-5} & low, medium, high, xhigh, max &
provider defaults, no sampling or length control sent\\
Sonnet-4.6 & \texttt{claude-sonnet-4-6} & low, medium, high, xhigh &
provider defaults, no sampling or length control sent\\
Haiku-4.5 & \texttt{claude-haiku-4-5} & low, medium, high, xhigh &
provider defaults, no sampling or length control sent\\
Gemini-3.1-Pro & \texttt{gemini-3.1-pro-preview} & low, medium, high &
temperature 0, max tokens 4{,}096 in the sweep, omitted in the ladder\\
Gemini-3.5-Flash & \texttt{gemini-3.5-flash} & minimal, low, medium, high &
temperature 0, max tokens 4{,}096 in the sweep, omitted in the ladder\\
Gemini-3.1-Flash-Lite & \texttt{gemini-3.1-flash-lite-preview} & minimal,
low, medium, high & temperature 0, max tokens 4{,}096 in the sweep, omitted in the
ladder\\
Qwen3.5-27B & \texttt{Qwen/Qwen3.5-27B} & non-thinking, thinking &
temperature 0, max tokens 4{,}096 non-thinking, omitted thinking\\
Qwen3.5-9B & \texttt{Qwen/Qwen3.5-9B} & non-thinking, thinking &
temperature 0, max tokens 4{,}096 non-thinking, omitted thinking\\
Qwen3.5-4B & \texttt{Qwen/Qwen3.5-4B} & non-thinking, thinking &
temperature 0, max tokens 4{,}096 non-thinking, omitted thinking\\
\bottomrule
\end{tabular*}}
\end{table}

\subsection{Crossover at every effort rung}
\label{app:effort-crossover}

For six configurations, stored raw and disclosed-boundary replies are available at
every acted-on
rung. Table~\ref{tab:effort-crossover} additionally lists Gemini-3.1-Flash-Lite,
whose four byte-identical rungs are excluded from the six-configuration crossover
count. Replaying each through both transports yields the cells plotted in
Figure~\ref{fig:effort-crossover}. No additional model call is made.

{\small
\setlength{\tabcolsep}{5.0pt}
\setlength{\LTleft}{\fill}
\setlength{\LTright}{\fill}
\begin{longtable}{llrrrrrrr}
\caption{Crossover at every measured rung. Cells are percentages and effects
are percentage points, defined as in Table~\ref{tab:crossover}. An asterisk marks
the descriptive masked-fragility rule. Gemini-3.1-Flash-Lite returns byte-identical
replies at all four settings and is retained only to document that the provider did
not expose a usable ladder.}\label{tab:effort-crossover}\\
\toprule
Configuration & Rung & $RR$ & $RN$ & $NR$ & $NN$ & Damage & Compensation & Matched gap\\
\midrule
\endfirsthead
\multicolumn{9}{l}{\small\itshape Table~\thetable\ continued from previous page}\\
\toprule
Configuration & Rung & $RR$ & $RN$ & $NR$ & $NN$ & Damage & Compensation & Matched gap\\
\midrule
\endhead
\midrule
\multicolumn{9}{r}{\small\itshape Continued on next page}\\
\endfoot
\bottomrule
\endlastfoot
GPT-5.6-sol & low & 98.2 & 28.6 & 53.6 & 91.1 & $-69.6$ & $+62.5$ & $-7.1$\\
GPT-5.6-sol & medium & 94.6 & 30.4 & 48.2 & 92.9 & $-64.3$ & $+62.5$ & $-1.8$$^{\ast}$\\
GPT-5.6-sol & high & 98.2 & 32.1 & 48.2 & 98.2 & $-66.1$ & $+66.1$ & $+0.0$$^{\ast}$\\
GPT-5.6-sol & xhigh & 98.2 & 30.4 & 51.8 & 98.2 & $-67.9$ & $+67.9$ & $+0.0$$^{\ast}$\\
\midrule
Opus-5 & low & 98.2 & 28.6 & 57.1 & 60.7 & $-69.6$ & $+32.1$ & $-37.5$\\
Opus-5 & medium & 100.0 & 32.1 & 41.1 & 91.1 & $-67.9$ & $+58.9$ & $-8.9$\\
Opus-5 & high & 96.4 & 30.4 & 42.9 & 96.4 & $-66.1$ & $+66.1$ & $+0.0$$^{\ast}$\\
Opus-5 & xhigh & 98.2 & 32.1 & 42.9 & 100.0 & $-66.1$ & $+67.9$ & $+1.8$$^{\ast}$\\
Opus-5 & max & 100.0 & 32.1 & 46.4 & 98.2 & $-67.9$ & $+66.1$ & $-1.8$$^{\ast}$\\
\midrule
Gemini-3.1-Pro & low & 100.0 & 25.0 & 32.1 & 91.1 & $-75.0$ & $+66.1$ & $-8.9$\\
Gemini-3.1-Pro & medium & 98.2 & 26.8 & 41.1 & 91.1 & $-71.4$ & $+64.3$ & $-7.1$\\
Gemini-3.1-Pro & high & 98.2 & 26.8 & 35.7 & 91.1 & $-71.4$ & $+64.3$ & $-7.1$\\
\midrule
Fable-5 & low & 100.0 & 30.4 & 48.2 & 71.4 & $-69.6$ & $+41.1$ & $-28.6$\\
Fable-5 & medium & 100.0 & 30.4 & 48.2 & 75.0 & $-69.6$ & $+44.6$ & $-25.0$\\
Fable-5 & high & 98.2 & 32.1 & 42.9 & 96.4 & $-66.1$ & $+64.3$ & $-1.8$$^{\ast}$\\
Fable-5 & xhigh & 100.0 & 33.9 & 46.4 & 92.9 & $-66.1$ & $+58.9$ & $-7.1$\\
Fable-5 & max & 100.0 & 30.4 & 35.7 & 100.0 & $-69.6$ & $+69.6$ & $+0.0$$^{\ast}$\\
\midrule
Gemini-3.5-Flash & minimal & 96.4 & 23.2 & 75.0 & 57.1 & $-73.2$ & $+33.9$ & $-39.3$\\
Gemini-3.5-Flash & low & 92.9 & 25.0 & 67.9 & 58.9 & $-67.9$ & $+33.9$ & $-33.9$\\
Gemini-3.5-Flash & medium & 96.4 & 23.2 & 76.8 & 64.3 & $-73.2$ & $+41.1$ & $-32.1$\\
Gemini-3.5-Flash & high & 96.4 & 23.2 & 71.4 & 62.5 & $-73.2$ & $+39.3$ & $-33.9$\\
\midrule
Opus-4.8 & low & 87.5 & 28.6 & 69.6 & 39.3 & $-58.9$ & $+10.7$ & $-48.2$\\
Opus-4.8 & medium & 92.9 & 28.6 & 67.9 & 48.2 & $-64.3$ & $+19.6$ & $-44.6$\\
Opus-4.8 & high & 96.4 & 33.9 & 64.3 & 50.0 & $-62.5$ & $+16.1$ & $-46.4$\\
Opus-4.8 & xhigh & 98.2 & 32.1 & 60.7 & 62.5 & $-66.1$ & $+30.4$ & $-35.7$\\
Opus-4.8 & max & 98.2 & 30.4 & 44.6 & 94.6 & $-67.9$ & $+64.3$ & $-3.6$$^{\ast}$\\
\midrule
Gemini-3.1-Flash-Lite & minimal & 78.6 & 19.6 & 80.4 & 14.3 & $-58.9$ & $-5.4$ & $-64.3$\\
Gemini-3.1-Flash-Lite & low & 78.6 & 19.6 & 80.4 & 14.3 & $-58.9$ & $-5.4$ & $-64.3$\\
Gemini-3.1-Flash-Lite & medium & 78.6 & 19.6 & 80.4 & 14.3 & $-58.9$ & $-5.4$ & $-64.3$\\
Gemini-3.1-Flash-Lite & high & 78.6 & 19.6 & 80.4 & 14.3 & $-58.9$ & $-5.4$ & $-64.3$\\
\end{longtable}
}

\section{Statistical Details}
\label{app:study-a-multiplicity}

The main text treats the 14 operation families as inferential units. For each
of the two primary Study-A components reported here, we apply Holm's step-down
procedure across the eight
model-specific enumerated sign-flip tests. Because the families are purposively
constructed, the $p$ values use a family-sign symmetry null: conditional on the
observed effect magnitudes, positive and negative signs are exchangeable. Enumeration
is exact for this finite family set.
Table~\ref{tab:holm} reports the two primary components with 95\% intervals from
a scenario-family percentile bootstrap of the mean (10{,}000 replicates,
families resampled as units). The transport-damage result also has a direct finite-benchmark reading: every
model-specific effect is negative and every leave-one-family-out range remains
negative. The largest adjusted $p$ is .001465.

\begin{table}[h]
\centering
\small\setlength{\tabcolsep}{4pt}
\caption{Enumerated and Holm-adjusted two-sided family-sign $p$ values for the two
primary Study-A components under the single-clause disclosed-boundary contract. Effect sizes are
percentage points.}
\label{tab:holm}
\begin{tabular}{lrrrrrr}
\toprule
& \multicolumn{3}{c}{Fixed-reply transport} &
\multicolumn{3}{c}{Contract-conditioned compensation}\\
Model & Effect [95\% CI] & Enum.\ $p$ & Holm $p$ & Effect [95\% CI] & Enum.\ $p$ & Holm $p$\\
\midrule
GPT-5.6-sol & $-64.3$ $[-80.4, -46.4]$ & .000244 & .001465 & $+60.7$ $[+46.4, +75.0]$ & .000244 & .001953\\
GPT-5.5 & $-71.4$ $[-85.7, -55.4]$ & .000244 & .001465 & $+60.7$ $[+44.6, +75.0]$ & .000244 & .001953\\
Opus-5 & $-66.1$ $[-82.1, -50.0]$ & .000244 & .001465 & $+58.9$ $[+41.1, +75.0]$ & .000488 & .002930\\
Gemini-3.1-Pro & $-73.2$ $[-87.5, -58.9]$ & .000122 & .000977 & $+55.4$ $[+33.9, +73.2]$ & .001221 & .006104\\
Gemini-3.5-Flash & $-67.9$ $[-82.1, -51.8]$ & .000244 & .001465 & $+30.4$ $[+12.5, +48.2]$ & .013672 & .041016\\
Opus-4.8 & $-64.3$ $[-80.4, -46.4]$ & .000244 & .001465 & $+30.4$ $[+14.3, +48.2]$ & .003906 & .015625\\
Qwen3.5-27B & $-55.4$ $[-69.6, -41.1]$ & .000244 & .001465 & $0.0$ $[0.0, 0.0]$ & 1.000000 & 1.000000\\
Gemini-3.1-Flash-Lite & $-58.9$ $[-71.4, -48.2]$ & .000122 & .000977 & $-5.4$ $[-10.7, 0.0]$ & .250000 & .500000\\
\bottomrule
\end{tabular}
\end{table}

A configuration joins the supported positive-compensation set when its effect is
positive, its Holm-adjusted $p$ is at most .05, and its leave-one-family-out
estimates stay positive. Six qualify: GPT-5.6-sol, GPT-5.5, Opus-5,
Gemini-3.1-Pro, Opus-4.8, and
Gemini-3.5-Flash. The supported set is defined over the eight same-window rows alone.

\subsection{Userland robustness}
\label{app:userland}

We report the pinned GNU/Linux replay; the BSD/macOS execution changes only the
utility environment, not the stored reply. For the six evaluated provider-hosted
configurations, corresponding crossover cells differ by at most 3.6 points. The two
remaining configurations differ by 7.1--12.5 points. Fixed-reply damage remains
negative for all eight configurations in both userlands, with largest Holm-adjusted
$p$ values of .001465 (GNU) and .001709 (BSD), and the masked set is unchanged. Four
of 48 ladder comparisons change their internal rung order, so we report one primary
userland.

The earlier BSD-live campaign provides a separate cross-userland transfer analysis.
Its raw and nested commands were elicited in BSD/macOS sessions and replayed unchanged
in the pinned GNU container. Table~\ref{tab:dialect-transfer} therefore measures how
BSD-elicited commands transfer across utility dialects; it does not estimate what the
same models would generate if explicitly prompted for GNU. The direction is
model-specific, the largest shift is 8.9 points, and both raw and nested rankings
change across userlands.

\begin{table}[H]
\centering
\small
\caption{Cross-userland transfer in the earlier BSD-live campaign. Each cell reports
pass rate on BSD live execution and GNU replay of the identical stored command. Arms
retain their original campaign multiplicity: raw cells contain 56 records, while
nested cells contain 56 or 168 depending on the configuration. We interpret only the
within-arm BSD$\to$GNU change. The final column summarizes the higher-transfer
userland; it is not a counterfactual GNU-targeted generation result.}
\label{tab:dialect-transfer}
\begin{tabular*}{\textwidth}{@{\extracolsep{\fill}}lccc@{}}
\toprule
Model & Raw BSD$\to$GNU & Nested BSD$\to$GNU & Transfers better to\\
\midrule
Fable-5 & $96.4\!\to\!91.1$ & $92.9\!\to\!87.5$ & BSD\\
Qwen3.5-27B (non-thinking) & $78.6\!\to\!87.5$ & $25.0\!\to\!32.1$ & GNU\\
Qwen3.5-27B (thinking) & $73.2\!\to\!82.1$ & $44.6\!\to\!50.0$ & GNU\\
Gemini-3.1-Pro & $92.9\!\to\!100.0$ & $89.3\!\to\!96.4$ & GNU\\
Gemini-3.5-Flash & $100.0\!\to\!96.4$ & $69.6\!\to\!67.9$ & BSD\\
Opus-4.8 & $91.1\!\to\!87.5$ & $73.8\!\to\!73.8$ & BSD (raw); tie (nested)\\
\bottomrule
\end{tabular*}
\end{table}

The raw ordering changes from Gemini-3.5-Flash/Fable-5/Gemini-3.1-Pro on BSD to
Gemini-3.1-Pro/Gemini-3.5-Flash/Fable-5 on GNU; the nested ordering likewise swaps
Fable-5 and Gemini-3.1-Pro at the top. Utility dialect is therefore a second systems
axis for measured shell competence, distinct from quoting reliability.

Study B contains the only effect-sign changes: Gemini-3.1-Pro and Fable-5 move from
small positive native-minus-raw effects on BSD to small negative effects under GNU
(Table~\ref{tab:native}). This comparison also transfers the same BSD-elicited
commands to GNU, keeping generation fixed.

\subsection{Masked fragility and the diagonal identity}
\label{app:masked-fragility}

As one illustrative reading aid (not a fitted criterion), a configuration may be
called descriptively masked when the matched gap is small while both components
are large, for instance
\[
|Y_{NN}-Y_{RR}|\leq5\text{ pp},\quad
Y_{RN}-Y_{RR}\leq-40\text{ pp},\quad
Y_{NN}-Y_{RN}\geq30\text{ pp}.
\]
At the reported rungs GPT-5.6-sol meets this reading ($-3.6 = -64.3 + 60.7$),
and ten of the 30 rung-level crossovers meet the same cut (asterisks in
Table~\ref{tab:effort-crossover}), all at ladder tops. Masking depends on the
operating point and the chosen cut, not on a model-level invariant.

This descriptive threshold identifies cancellation between two large components. The decomposition is an identity verified for every task before
aggregation:
\[
Y_{NN}-Y_{RR}=(Y_{RN}-Y_{RR})+(Y_{NN}-Y_{RN}).
\]

\section{Native-Contract Diagnostics}
\label{app:family-detail}

Study B contains 8,736 generated arm records: 4,368 raw and 4,368 native.
Each stored generation is replayed in both BSD and GNU userlands, yielding 17,472
execution outcomes. The main text reports aggregate GNU pass rates. This appendix adds the paired effects, robustness comparisons, and failure diagnostics.

Table~\ref{tab:native} reports the paired contract effects. $\Delta$ is the
campaign-average native-minus-raw change in percentage points. LOFO gives the range when each operation family is omitted in turn. PF/FP counts paired pass$\to$fail and fail$\to$pass transitions. The BSD columns provide the userland robustness
comparison of Appendix~\ref{app:userland}.

\begin{table}[h]
\centering
\small
\caption{Exploratory native-minus-raw effects over each measured effort ladder,
ordered by GNU replay effect.}
\label{tab:native}
{\setlength{\tabcolsep}{6.2pt}
\begin{tabular}{lccc@{\hspace{1.0em}}ccc}
\toprule
& \multicolumn{3}{c}{\textbf{BSD live}} & \multicolumn{3}{c}{\textbf{GNU replay}}\\
\cmidrule(lr){2-4}\cmidrule(lr){5-7}
Model & $\Delta$ & LOFO & PF/FP & $\Delta$ & LOFO & PF/FP\\
\midrule
Opus-4.8 & $+3.10$ & $[+2.18,+3.46]$ & 5/31 & $+2.62$ & $[+1.67,+2.95]$ & 11/33\\
Opus-5 & $-0.36$ & $[-0.51,0.00]$ & 12/9 & $-0.83$ & $[-1.03,-0.51]$ & 16/9\\
Fable-5 & $+0.95$ & $[0.00,+1.15]$ & 3/11 & $-2.14$ & $[-2.44,-0.64]$ & 22/4\\
GPT-5.6-sol & $-1.19$ & $[-1.92,-0.48]$ & 25/17 & $-2.53$ & $[-3.37,-0.64]$ & 26/9\\
Gemini-3.1-Pro & $+1.98$ & $[-0.64,+3.21]$ & 8/18 & $-3.77$ & $[-4.91,-0.64]$ & 23/4\\
Gemini-3.5-Flash & $-4.02$ & $[-4.33,-3.04]$ & 37/10 & $-9.97$ & $[-10.74,-6.89]$ & 77/10\\
\bottomrule
\end{tabular}}
\end{table}

Process exit status misses a substantial share of failures. Across the four
contract--userland conditions, 41--62 executions exit zero while leaving the wrong
state, accounting for 23.4--47.0\% of that condition's failures
(Table~\ref{tab:failure-taxonomy}). Native one-call schema adherence is
98.2--100\%.

\begin{table}[htbp]
\centering
\small
\caption{Study-B execution outcomes, aggregated across six models. Each row
partitions 4,368 executions. Adherence denotes an invalid one-call tool invocation. Syntax includes parser and command-usage errors. Exit-0 wrong is a silent final-state failure.}
\label{tab:failure-taxonomy}
{\setlength{\tabcolsep}{7.0pt}
\begin{tabular}{llrrrrr}
\toprule
Userland & Contract & Pass & Adherence & Syntax & Nonzero & Exit-0 wrong\\
\midrule
BSD & raw & 4230 & 0 & 31 & 59 & \textbf{48}\\
BSD & native & 4236 & 20 & 9 & 41 & \textbf{62}\\
GNU & raw & 4252 & 0 & 39 & 36 & \textbf{41}\\
GNU & native & 4146 & 20 & 22 & 128 & \textbf{52}\\
\bottomrule
\end{tabular}}
\end{table}

\subsection{Study-B robustness and failure analysis}
\label{app:native-inference}

The family-sign sensitivity analysis treats the 14 operation families as the
inferential units and applies Holm adjustment across the six models separately in
each userland. All twelve adjusted values exceed .05. The minimum is .18750 for Gemini-3.5-Flash in both userlands. Study B is therefore descriptive and
exploratory. Table~\ref{tab:native} reports the effect sizes, leave-one-family-out
ranges, and paired transitions that support that interpretation.

\subsubsection{Effects by effort rung}
\label{app:native-effort}

\begin{figure}[h]
\centering
\includegraphics[width=\linewidth]{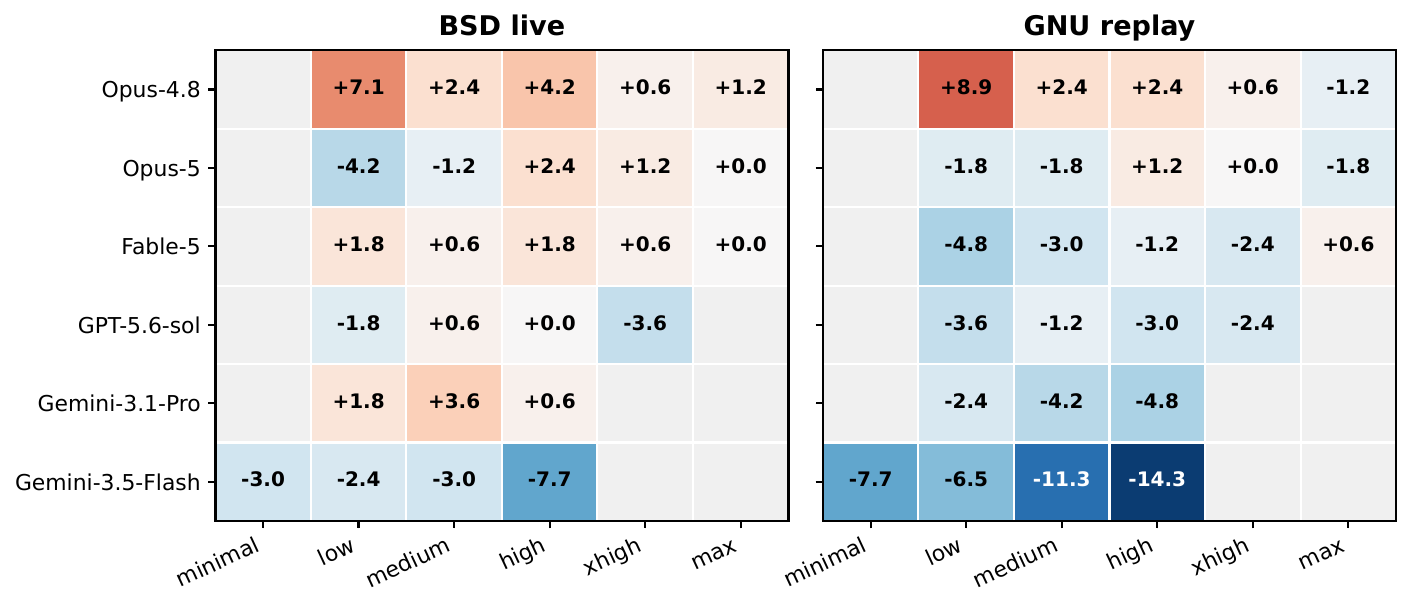}
\caption{Native-minus-raw pass-rate change by provider effort rung, in percentage
points on a zero-centered scale. Blank cells mark rungs the provider does not expose. Models follow descending GNU replay campaign effect. Labels are not
comparable compute budgets across providers.}
\label{fig:native-effort-heatmap}
\end{figure}

Figure~\ref{fig:native-effort-heatmap} decomposes the aggregate effects from the
main text. Each cell compares raw and native arms within one provider-specific
effort rung.

\paragraph{Wire integrity.}
Three model configurations from two providers retained raw native-tool arguments,
allowing byte-level verification
between the decoded \texttt{command} field and executor input: 660 records for
Gemini-3.5-Flash, 497 for Gemini-3.1-Pro, and 672 for GPT-5.6-sol. Nineteen adherence
failures contain no usable argument. For Opus-4.8, Opus-5, and Fable-5, the stored
artifact begins at the decoded command string, so their analysis starts at that
boundary.

\section{Replication and Transport Robustness}
\label{app:heldout}

\paragraph{Design.}
The private-v2 single-clause crossover uses 42 unpublished hostile payloads: three variants for each of
the 14 operation families. The task manifest, contract order, and sample hash were
fixed before inference. GPT-5.6-sol and Opus-4.8 each contribute one accepted reply
per task under the raw and single-clause disclosed-boundary contracts, for 168
accepted generations. Raw- and disclosed-boundary-contract calls were interleaved
within one serving window. Every
stored reply was replayed through both transports and scored by the same final-state
validators used for the public core. Table~\ref{tab:heldout} reports the
resulting effects.

Here and in the private-v1 campaign, an accepted reply is the first usable response
returned for a scheduled model--task--contract cell. Semantic failure, truncation,
refusal, malformed output, and contract error are retained as outcomes rather than
retried or filtered. At the campaign-invocation level, private-v2 made 174 attempts:
168 returned replies were retained, six provider or transport errors that yielded no
usable response were retried, and no returned replies were rejected. Private-v1 made
253 attempts across three configurations and two contracts: 252 returned replies
were retained, one no-response error was retried, and no returned replies were
rejected. Lower-level retries inside provider adapters are not separately observable.

\paragraph{Validator checks.}
The 42 private tasks accept their oracles under both transports and reject every
untouched initial state. Naive probes produce no false accepts, and mutation testing
rejects all 148 applicable corruptions. These checks cover the declared corruption
classes.

\paragraph{Scope.}
The private set contains hostile payloads only and was designed to test whether the
mechanism transfers to unseen literals. Its difficulty was not matched to the
56-task public core, so absolute rates should be read within the private set. The
result establishes transfer of fixed-reply transport damage across payload samples. Repeated-generation stability, temporary-script bypass, and
typed-operation evidence are reported in Appendix~\ref{app:robustness-detail}.

\subsection{Bypass, repeated draws, and typed operations}
\label{app:robustness-detail}

\subsubsection{Temporary-script bypass}

A public-core replication executes every stored trial-0 raw reply of the eight
same-window configurations through raw \texttt{bash -c}, the nested wrapper, and a
temporary Bash script. The script transport reproduces the raw-path outcome for
all 448 configuration--task pairs, recovering all 292 nested-only failures and
rescuing no raw-path failure. An escaped variant of the nested transport, which
interpolates each stored reply with standard shell quoting instead of verbatim
substitution, reproduces the raw-path outcome for the same 448 pairs: correct
escaping at the boundary removes the entire effect. The released records
reproduce both replays without private data.

Table~\ref{tab:script-bypass} uses the earlier private-v1 raw-generation campaign,
which includes Qwen3.5-27B and is distinct from the private-v2 generations in
Table~\ref{tab:heldout}. Each accepted raw-contract reply runs through three paths:
raw \texttt{bash -c}, the nested wrapper, and a temporary Bash script. The fixed
reply makes the comparison a direct test of transport behavior.
Table~\ref{tab:script-bypass} reports the resulting pass counts.

\subsubsection{Advice arm}
\label{app:advice-arm}

Table~\ref{tab:advice-arm} reports matched nested success under the advice-free
and advice-bearing disclosed contracts, collected for each configuration in one
serving window and replayed with the public validators. $\Delta$ is the paired
same-window difference, free of the cross-window drift the repeated draws bound
(Appendix~\ref{app:repeat-generation}).

\begin{table}[h]
\centering
\small
\caption{Matched nested success (of 56) under the disclosed contract without and
with escaping advice, paired within one serving window. Advice barely moves the
top of the ladder, moves the middle of the ladder the most, and does not reach the
bottom.}
\label{tab:advice-arm}
\begin{tabular}{lrrr}
\toprule
Configuration & Advice-free & With advice & $\Delta$ (pp)\\
\midrule
GPT-5.6-sol & 54 & 49 & $-8.9$\\
GPT-5.5 & 49 & 53 & $+7.1$\\
Opus-5 & 52 & 54 & $+3.6$\\
Sonnet-4.6 & 32 & 46 & $+25.0$\\
Haiku-4.5 & 25 & 32 & $+12.5$\\
Opus-4.8 & 30 & 34 & $+7.1$\\
Qwen3.5-27B & 17 & 18 & $+1.8$\\
Gemini-3.1-Flash-Lite & 7 & 8 & $+1.8$\\
\bottomrule
\end{tabular}
\end{table}

\subsubsection{Grammar crossover}
\label{app:grammar-crossover}

Table~\ref{tab:grammar-crossover} crosses the disclosed grammar (double- or
single-quote wrapper) with the executed wrapper. Each disclosed reply is replayed
on both the wrapper it was told about and the other one; a model that adapts to
the stated grammar passes more on the matched diagonal than the mismatched
anti-diagonal. All executions are zero-call replays.

\begin{table}[h]
\centering
\small
\caption{Grammar crossover (tasks passed of 56). Rows are the disclosed grammar,
columns the executed wrapper; the diagonal is matched, the anti-diagonal
mismatched. The diagonal advantage separates the same top, middle, and bottom
groups as the matched-nested scores.}
\label{tab:grammar-crossover}
\begin{tabular}{lrrrrr}
\toprule
& \multicolumn{2}{c}{Double disclosed} & \multicolumn{2}{c}{Single disclosed} & Diag$-$anti\\
Configuration & on double & on single & on single & on double & (pp)\\
\midrule
GPT-5.6-sol & 54 & 7 & 53 & 10 & $+80.4$\\
GPT-5.5 & 49 & 8 & 54 & 8 & $+77.7$\\
Opus-5 & 52 & 8 & 45 & 16 & $+65.2$\\
Sonnet-4.6 & 32 & 11 & 23 & 16 & $+25.0$\\
Opus-4.8 & 30 & 11 & 23 & 18 & $+21.4$\\
Haiku-4.5 & 25 & 13 & 20 & 11 & $+18.8$\\
Qwen3.5-27B & 17 & 11 & 11 & 17 & $+0.0$\\
Gemini-3.1-Flash-Lite & 7 & 19 & 9 & 19 & $-19.6$\\
\bottomrule
\end{tabular}
\end{table}

\subsubsection{Real-\texttt{ssh} full crossover}
\label{app:ssh-full}

Table~\ref{tab:ssh-full} completes the real-\texttt{ssh} 2$\times$2 by replaying
the disclosed-boundary replies as well, giving $NR$ and $NN$ on the real path.
Compensation ($NN-RN$) on the real boundary matches the synthetic nested
compensation exactly for five of six replayed configurations;
Gemini-3.1-Flash-Lite differs by one task ($-1.8$ versus $-5.4$ points). The two
non-adapting configurations show no positive compensation on either path. Zero
model calls.

\begin{table}[h]
\centering
\small
\caption{Real-\texttt{ssh} full crossover (tasks passed of 56). Damage is $RN-RR$
and compensation is $NN-RN$, both on the real \texttt{ssh} path. The final column
repeats the synthetic nested compensation for comparison.}
\label{tab:ssh-full}
\begin{tabular}{lrrrrrr}
\toprule
Configuration & RR & RN & NR & NN & \texttt{ssh} comp. & Nested comp.\\
\midrule
GPT-5.6-sol & 53 & 17 & 31 & 51 & $+60.7$ & $+60.7$\\
GPT-5.5 & 56 & 16 & 28 & 50 & $+60.7$ & $+60.7$\\
Opus-5 & 54 & 17 & 24 & 50 & $+58.9$ & $+58.9$\\
Opus-4.8 & 51 & 15 & 35 & 32 & $+30.4$ & $+30.4$\\
Qwen3.5-27B & 48 & 17 & 47 & 17 & $+0.0$ & $+0.0$\\
Gemini-3.1-Flash-Lite & 44 & 11 & 45 & 10 & $-1.8$ & $-5.4$\\
\bottomrule
\end{tabular}
\end{table}

\subsubsection{JSON serializer boundary}
\label{app:json-boundary}

Table~\ref{tab:json-boundary} replays each stored raw reply through a JSON
tool-call boundary two ways: a correct serializer (\texttt{json.dumps} then
\texttt{json.loads}) that round-trips the reply, and a naive embedding that
pastes the reply into a JSON string field without escaping. The naive boundary
re-parses the reply's double quotes and backslashes as JSON syntax and breaks on
the same characters as the shell boundary. Many replies do not even parse, so
its damage is comparable to the shell nested transport, while the correct
serializer costs nothing. Zero model calls.

\begin{table}[h]
\centering
\small
\caption{JSON serializer boundary (tasks passed of 56). Correct-serializer damage
is $\approx 0$; naive-embedding damage is comparable to the shell nested
transport. \emph{Unparseable} counts replies whose naive JSON embedding fails to
parse.}
\label{tab:json-boundary}
\begin{tabular}{lrrrr}
\toprule
Configuration & Raw & Correct & Naive & Naive damage\\
\midrule
GPT-5.6-sol & 53 & 53 & 21 & $-57.1$\\
GPT-5.5 & 56 & 56 & 19 & $-66.1$\\
Opus-5 & 54 & 54 & 22 & $-57.1$\\
Opus-4.8 & 51 & 51 & 18 & $-58.9$\\
Qwen3.5-27B & 48 & 48 & 19 & $-51.8$\\
Gemini-3.1-Flash-Lite & 44 & 44 & 8 & $-64.3$\\
\bottomrule
\end{tabular}
\end{table}

\subsubsection{Repeated generations}
\label{app:repeat-generation}

A public-core replication adds two generations per contract for all eight
same-window configurations. Across the three draws, damage stays negative for
every configuration and draw, with per-configuration ranges of 1.8--7.1 points.
No compensation changes sign: the largest spread is 12.5 points (Opus-4.8), and
Qwen3.5-27B realizes exactly zero compensation in every draw. Scoring each configuration by the tasks it passes in all three draws
still yields five strict rank reversals between the matched contracts. In the
trial-0 draws, 26 of 28 pairs are strictly comparable (non-tied under both matched
contracts) and five reverse: GPT-5.6-sol versus Gemini-3.5-Flash, Gemini-3.1-Pro,
GPT-5.5, and Opus-5, and Opus-5 versus Gemini-3.1-Pro. Consistent with the main text, the one reversal that is unambiguous at this
resolution is GPT-5.6-sol versus Gemini-3.5-Flash (behind by one task under $RR$,
ahead by eighteen under $NN$); the other four rest on a single-task margin on at
least one side. The same five reverse under the all-draw criterion.

On the private payloads, we collect three additional generations for eight tasks under both
generation contracts
and both models. Across the four draws, GPT-5.6-sol shows mean damage of $-78.1$ points and mean compensation of $+81.3$. The corresponding values for Opus-4.8 are $-84.4$ and $+46.9$. Reply text varies in 9/16 task--contract cells for GPT-5.6-sol
and 10/16 for Opus-4.8, yet every draw preserves negative damage and positive
compensation. The mechanism therefore persists across distinct sampled replies.

\subsubsection{Typed operations}

A representation study replaces shell-string construction with structured operations
on 18 private tasks from six naturally typeable families. Across two models, typed
operations pass 25/36 tasks (69.4\%), compared with 35/36 (97.2\%) for raw
\texttt{bash -c} and temporary scripts. Of the eleven typed failures, ten reach the
wrong final state and one fails during execution. Structured actions remove one
quoting surface, while literal-preservation errors remain in arguments and payload
fields. The two configurations probe representation sensitivity.

\section{Validation and Reproduction}
\subsection{Validator mutation audit}
\label{app:validator-audit}

Validator checks begin from each oracle-produced valid state, apply every applicable
mutation class, and rerun the validator:

\begin{center}
\begin{tabular}{lr}
\toprule
Mutation class & Rejected\\
\midrule
Delete a changed/required file & 60/60\\
Flip one byte in a changed file & 60/60\\
Insert an unexpected collateral file & 56/56\\
Restore a file that should be removed & 17/17\\
Amend a Git-only final state & 4/4\\
\midrule
Total & 197/197\\
\bottomrule
\end{tabular}
\end{center}

The validators accept every oracle and benign tier-0 probe. They reject every
untouched fixture, hostile probe, and all 197 mutated states. These checks cover the enumerated invalid states. Unenumerated false positives remain possible.

\subsection{Execution and reproduction}
\label{app:reproduction}

Each command runs in a fresh fixture with a trimmed environment and a 15-second
timeout. The local runner invokes \texttt{bash -c} through \texttt{execve}. The reported GNU replay uses a pinned, network-disabled container. All 56 oracles must
pass in that container before crossover replay, and replay never queries a model.

The public code artifact is available at
\url{https://github.com/LeonardNJU/quoteBench}. It contains the provider-agnostic
harness, all 56 public tasks, validators, contract prompts, and offline rollout
verification and descriptive-analysis commands. Together with the separately hosted
rollout archive and its SHA-256 manifest, these files reproduce the released
campaign/model/contract/effort/trial/toolchain rates and the public GNU crossover
table. \texttt{REPRODUCE.md} documents the commands, package layout, and
campaign-level count reconciliation. Serving-path, authentication, private-payload,
and internal adapter metadata are excluded.

Private-payload records are withheld to preserve held-out evaluation and are not part
of the public release. They are used only for the private replication and mitigation
analyses reported in this paper; neither the payloads nor replies are included in the
arXiv source package or ancillary files.

\paragraph{Rollout archive.}
The sanitized public rollout archive at
\url{https://huggingface.co/datasets/lsamc/QuoteBench-Rollouts} contains
12{,}999 records across 33 arm files in the
\texttt{quotebench-rollout-v1} schema; each record carries one generation with
its replays, prompt, reply, identifiers, usage, and final-state outcomes.

\end{document}